\documentclass[11pt]{article}

\usepackage[a4paper,margin=25mm]{geometry}
\usepackage[T1]{fontenc}
\usepackage{lmodern}
\usepackage{graphicx}
\usepackage{booktabs}
\usepackage{amsmath}
\usepackage{siunitx}
\usepackage{caption}
\usepackage[hidelinks]{hyperref}
\usepackage{authblk}

\title{\bfseries What Input Resolution Is Required for Bird Species Identification,\\
and What Is Its Latency Cost on an Edge Device?\\[2mm]
\large A Study of 14 Input Resolutions and Six Architectures\\ with On-Device Measurements}

\author[1]{Takeshi Nishikawa}
\affil[1]{Foundation for Computational Science (FOCUS), Kobe, Japan\\
\texttt{nishikawa@j-focus.or.jp}}
\date{2026}

\begin{document}
\maketitle

\begin{abstract}
\noindent
Bird-strike mitigation at wind farms requires identifying distant birds that may occupy image regions only a few tens of pixels wide. The design problem therefore reduces to one question: which input resolution $N$ gives high identification accuracy at the shortest processing time, and what are that accuracy and time? We answer it with a factorial design over 14 input side lengths $N$ from 16 to 224 pixels, six architectures, two training and evaluation regimes, and 30 random seeds---2{,}520 trained models and 5{,}040 evaluation conditions in all---combined with latency measurements on an NVIDIA Jetson Orin Nano and an H200 GPU. We report four findings. First, the selected configuration depends on the accuracy target, the candidate set, the preprocessing path and the precision assignment; configurations are selected on validation and reported on test. Among the 84 base candidates, a validation target of 0.90 selects ResNet50 at $N{=}112$, with a test accuracy of 0.8980 and an estimated classification time of \SI{1.85}{\milli\second}, and a target of 0.95 selects DINOv2-L at $N{=}144$ (test 0.9630, \SI{12.70}{\milli\second}); admitting 14 further ViT-S/16 FP32 configurations changes the selection over the 0.931--0.938 band and under the \SI{10}{\milli\second} budget. For the 0.90 target on the base candidate set the operating point is $N\approx112$, and raising $N$ further yields little accuracy while increasing processing time. Changing the model raises accuracy more than raising the resolution ($+5.93$ percentage points from ResNet50 to DINOv2-L at the same $N{=}112$, versus $+2.33$ points from raising $N$ from 112 to 224), but the two levers cost differently in time ($7.7\times$ against $1.66\times$), so this is not a comparison at equal cost. The benefit of \emph{lowering} $N$, by contrast, differs substantially between the preprocessing paths one might assume: on the offline evaluation path, where every individual is decoded from its own image file, resolution-independent decoding imposes a floor and $N{=}224\rightarrow80$ saves only 13.5\%, whereas on the deployment path, where the 4K frame is decoded once by the detector and the classifier only crops and resizes, the same change saves 46.2\%. Second, retraining at each input resolution has architecture-dependent effects: it improves CNN accuracy by 15--48 percentage points at $N\le48$, but reduces accuracy for the self-supervised ViT-L models at most evaluated resolutions above $N{=}16$; the direction is consistent across all 30 seeds for the CNNs at $N{=}16$--$80$ and for the two ViT-L models at $N{=}32$--$64$. Third, although the \emph{total} input pixel count differs by a factor of 196 between $N{=}16$ and $N{=}224$, Orin GPU latency differs by only 1.48--2.71$\times$ across all six models; the value of low-resolution accuracy therefore lies in relaxed sensor and optical requirements as well as in speed. Fourth, ViT-L-scale models fit this edge device even as a single unsplit engine, but activation magnitudes reach $3.6\times10^{5}$ and exceed the FP16 limit of 65{,}504, so with the build procedure and engine configuration used here an unsplit build had to run entirely in FP32; built in FP16 it returned output invariant to its input. Splitting the graph at transformer-block boundaries confines FP32 to the affected segments, which makes the deployed DINOv2-L chain 1.85$\times$ faster than the single-engine build (\SI{8.657}{} versus \SI{16.023}{\milli\second} at $N{=}32$).
\end{abstract}

\section{Introduction}

Wind-farm bird-strike mitigation systems must detect a bird, identify its species, and decide whether to curtail the turbine while the bird is still far away~\cite{marques}. At distances of approximately \SI{1}{\kilo\meter}, a target bird may occupy an image region only a few tens of pixels wide.

One unit of the system we target consists of three perimeter cameras and one tracking PTZ camera---four 4K (3840$\times$1920) streams---time-shared by a single Jetson Orin Nano 8GB, and two such units (eight cameras, two Orins) provide 360-degree coverage; this is the configuration of~\cite{v6}. A single camera can run at up to 30 fps, but the minimum requirement for a bird-strike prediction and avoidance system is to process all four streams of a unit at 10 fps. In operation the system extracts, from among many tracked objects, those likely to enter the rotor swept area within \SI{60}{\second}; it then classifies their species and decides whether to emit a deterrent sound or signal the turbine to stop. Each individual is classified once while it remains identified as the same track, with further classifications when it is predicted to reach two or three rotor radii of the turbine 3, 2 and 1 minutes ahead---at most four in total. How many individuals a single camera's field of view can handle is therefore also a performance metric, which we quantify in \S\ref{sec:optimal}.

The practical design question is therefore not the maximum accuracy a classifier can reach, but:

\begin{quote}
\emph{Which input side length $N$ gives high identification accuracy at the shortest processing time, and what are that accuracy and time?}
\end{quote}

The premises of this task differ from those of the usual accuracy/latency trade-off. In the usual framing, input resolution is a fixed specification and model complexity is reduced to improve latency. Here the input side length per individual is itself a design variable set by the optics: if the system tolerates a smaller $N$, processing time is reduced \emph{and} the same camera can identify birds at longer range. The value of preserving accuracy at low resolution therefore lies both in latency and in \emph{relaxed requirements on the optics and sensor}. The latter is decisive, since it determines whether a specification is physically attainable (\S\ref{sec:optics}).

The dependence of accuracy on resolution has been studied as a train/test discrepancy~\cite{touvron}, but three things matter for edge deployment and are not settled: the accuracy attainable when models are \emph{retrained} per resolution, the measured relationship between resolution and inference time on the target device, and how the required input side length differs across architecture families. We measure all three under a common dataset, training protocol and seed configuration.

\paragraph{Contributions.}
\begin{enumerate}
\item \textbf{$N\approx112$ is the operating point for a target accuracy of 0.90 on the base candidate set, and the selection moves with the target, the candidate set, the preprocessing path and the precision assignment} (\S\ref{sec:optimal}). Configurations are selected on validation and reported on test. Among the 84 base candidates, a validation target of 0.90 selects ResNet50 at $N{=}112$ (test 0.8980) at an estimated classification time of \SI{1.85}{\milli\second} on the deployment path, 0.93 selects ResNet50 at $N{=}224$ (test 0.9213, \SI{3.08}{\milli\second}) and 0.95 selects DINOv2-L at $N{=}144$ (test 0.9630, \SI{12.70}{\milli\second}); admitting 14 further ViT-S/16 FP32 configurations moves the recommendation over the 0.931--0.938 band and under the \SI{10}{\milli\second} budget. Raising $N$ beyond the selected point yields little accuracy while increasing processing time. Changing the model raises accuracy more than raising $N$ ($+5.93$ versus $+2.33$ percentage points), but costs $7.7\times$ the time against $1.66\times$, so the two levers are not compared at equal cost.
\item \textbf{Retraining at each input resolution has opposite effects across architecture families} (\S\ref{sec:adapt}). CNNs recover substantially when retrained at low resolution, whereas the self-supervised ViT-L models lose accuracy at most resolutions above $N{=}16$. The direction is consistent across all 30 seeds for the three CNNs at $N{=}16$--$80$ and for the two ViT-L models at $N{=}32$--$64$; at $N{=}80$ the ViT-L direction holds for 28 of 30 seeds for DINOv2-L and 29 of 30 for DINOv3-L, and from $N{=}96$ upward only 14--30 of 30 agree. These counts describe sign stability within the observed seeds and are not a test of whether the mean difference is zero.
\item \textbf{The reduction in processing time differs substantially between the assumed preprocessing paths} (\S\ref{sec:speed}). On GPU compute alone, a 196-fold difference in total input pixel count changes Orin latency by $1.48$--$2.71\times$. Defining the ratio as $t_{224}/t_{16}$, all six models lie in that band, including the two ViT-L-scale models (DINOv2-L $2.64$, DINOv3-L $2.71$); the same DINOv2-L on an H200 shows $1.45\times$. Once preprocessing is included, however, the verdict depends on which path is assumed. On the offline evaluation path, the estimated classification time drops only 13.5\% from $N{=}224$ to $N{=}80$; on the deployment path it drops 46.2\%, because the per-individual cost falls to \SI{0.136}{}--\SI{1.774}{\milli\second} and is dominated by $O(N^2)$ terms.
\item \textbf{Splitting ViT-L provides granularity of precision control, not feasibility} (\S\ref{sec:vitl}). The unsplit model already fits the device, but in the builds we produced its outlier activations forced the whole graph to FP32. Splitting at block boundaries confines FP32 to the affected intervals and leaves the rest in FP16, which is $1.85\times$ faster for DINOv2-L. An unsplit FP16 build instead returns a constant: 1{,}881 of 1{,}882 test images share identical logits.
\end{enumerate}

\section{Related work}

The dependence of accuracy on input resolution is classically framed as a train/test discrepancy: Touvron et al.~\cite{touvron} showed that fine-tuning at the evaluation resolution recovers much of the loss. We extend this analysis to the edge-deployment setting and show two effects that framing does not predict: the recovery has \emph{opposite signs} for CNNs and for self-supervised ViT-L models, and how much of the recovered accuracy translates into a latency benefit depends on which preprocessing path is assumed---on the deployment path lowering $N$ from 224 to 80 saves 46.2\%, on the offline path only 13.5\%. What we compare are two deployable pipelines rather than a single manipulated factor (\S\ref{sec:design}), so these effects are properties of that pair of pipelines and are not, on their own, evidence about fine-tuning in isolation.

Efficiency at the edge is usually pursued through lightweight architectures~\cite{qin,tan,he} and knowledge distillation~\cite{hinton}. We place these on a common footing and, in \S\ref{sec:kd}, isolate the contribution of distillation by varying only the loss. On the representation side, self-supervised foundation models~\cite{oquab,simeoni} are attractive because they retain discriminative power at small input sizes; we quantify their measured deployment cost.

That benchmark comparisons themselves can be governed by variance from seeds, data splits and the training pipeline was analysed by Bouthillier et al.~\cite{bouthillier}, who decompose the sources of variation. Our contribution is to carry that concern into a concrete deployment decision---which resolution, which architecture and which arithmetic precision to ship---and to quantify, on one candidate set, how much each evaluation choice moves the decision.

Extreme outliers in the activations of large transformers (``massive activations'') are known in language models and are a central obstacle to quantization~\cite{sun}. We show the same phenomenon breaks half-precision inference for vision foundation models at the edge, and that restricting FP32 to the affected interval is sufficient to avoid it.

\section{Experimental design}
\label{sec:design}

\paragraph{Design.} We used a factorial design with four factors. \emph{Resolution}: $N \in \{16, 32, \ldots, 224\}$, i.e.\ 14 levels in steps of 16 pixels.
\emph{Architecture}: MobileNetV4~\cite{qin}, EfficientNet-B0~\cite{tan}, ResNet50~\cite{he}, ViT-S/16~\cite{dosovitskiy},
DINOv2-L/14~\cite{oquab} and DINOv3-L/16~\cite{simeoni}.
\emph{Training regime}: A (degraded input to a 224-trained model) and B (resolution-native retraining).
\emph{Seed}: 42--71, i.e.\ 30 seeds (the distillation study of \S\ref{sec:kd} is the one exception and uses three).
The primary task is six-way raptor species identification of Japanese raptors (golden eagle, white-tailed eagle, Steller's sea eagle, mountain hawk-eagle, northern goshawk and eastern marsh harrier).
The dataset holds 12{,}519 images---4{,}918 from iNaturalist, 4{,}920 collected previously by our group and 1{,}000 extracted from video, with the remainder from other sources---split image-wise and stratified into 8{,}761 train, 1{,}876 validation and 1{,}882 test images, with the split held fixed across seeds.
Optimisation is AdamW (weight decay 0.05, label smoothing 0.1) with separate backbone and head learning rates ($5\times10^{-4}$/$5\times10^{-3}$ for the CNNs, $1\times10^{-4}$/$1\times10^{-3}$ for ViT-S/16, $5\times10^{-5}$/$5\times10^{-4}$ for the ViT-L-scale models). The epoch budget is 160 with batch 256 for the CNNs and ViT-S/16 and 60 with batch 64 and layer-wise decay 0.75 for the ViT-L-scale models; the weights of the best validation epoch are kept, with early stopping after 20 (CNN), 25 (ViT-S/16) or 10 (ViT-L) epochs without improvement.
Seeds vary initialisation, augmentation and shuffling only.

\paragraph{Definition of the input side length $N$.} We separate information content from input geometry. $N$ denotes the side length in pixels of the square input, so the total input pixel count is $N^{2}$. Every image passes $\mathrm{Resize}(256) \rightarrow \mathrm{CenterCrop}(224)$ and is then downscaled bicubically to $N \times N$; this is the step at which information is lost, and it is identical in both regimes. Regime~A then upsamples back to $224 \times 224$ before the model. Regime~B feeds $N \times N$ directly, except for DINOv2-L, whose patch-14 stem requires a multiple of 14: there the $N \times N$ image is resampled to the nearest multiple $N' \times N'$ (e.g.\ $N{=}64 \rightarrow N'{=}70$, an increase, or $N{=}32 \rightarrow N'{=}28$, a decrease) as the model input. Because $N'$ can exceed $N$, this must be kept in mind when comparing latencies across models.

\paragraph{What the comparison of the two regimes isolates.} Regimes A and B differ jointly in training resolution, learned weights, input geometry, and token or feature-map structure; for DINOv2-L they differ in the resampling to $N'$ as well. Their difference is therefore a comparison of two pipelines, not an isolated causal effect of retraining, and we report it as such throughout.

\paragraph{Scale.} We trained 2{,}520 models: 6 architectures $\times$ 14 resolutions $\times$ 30 seeds, with zero failed runs. These provide regime~B directly, while regime~A evaluates the $N{=}224$ models of each architecture and seed at every resolution, giving 5{,}040 evaluation conditions in total. On-device measurements cover 56 Orin configurations, 84 H200 configurations and 28 split ViT-L chains (14 resolutions $\times$ two models; 126 parts in total).

\paragraph{Measurement protocol.} The Orin Nano ran JetPack 6.2 (L4T R36.4.3) with TensorRT 10.3.0, in its default \texttt{MAXN\_SUPER} power mode (CPU \SI{1.73}{\giga\hertz}, GPU \SI{1020}{\mega\hertz}); the \SI{15}{\watt} mode was measured with the identical procedure and is compared in \S\ref{sec:power}. Every configuration in this \emph{GPU benchmark} is reported as the \emph{median of 30 runs}. A single run is \texttt{--warmUp=2000 --iterations=300 --avgRuns=100}, and we repeat it 30 times from a fresh process; the split ViT-L chains are measured the same way per part (\texttt{--iterations=200 --avgRuns=50}) and reported as the sum of the per-part medians over 126 parts. Across the 56 CNN and ViT-S/16 configurations the coefficient of variation was $0.12$--$0.59\%$ (median $0.23\%$). The other measurements follow their own designs and are not medians of 30 runs: CPU preprocessing ($n=200$ per configuration), the wall-clock of a single classification and the integrated detector throughput each use the sampling design stated where they appear (\S\ref{sec:speed}, \S\ref{sec:limits}). The H200 measurements used TensorRT 10.7 and are medians of three runs, so the repetition count is not matched to the Orin. Because batch-1 CNNs on H200 take only \SI{0.4}{\milli\second} and 32 of 84 configurations varied by more than 25\% across repeats, we do not base any claim on H200 CNN numbers.

\paragraph{Reference predictions.} Agreement rates compare an Orin engine against server FP32 predictions produced by \emph{the same checkpoint} that was exported to ONNX (seed 42). Comparing against a different checkpoint of the same architecture makes seed-to-seed disagreement appear as an effect of reduced precision.

\paragraph{Operating point.} As stated in \S1, one unit time-shares four 4K cameras on a single Jetson Orin Nano 8GB. Detection runs on every frame, so the minimum requirement of four streams at 10 fps (40 frames per second) leaves \SI{25.0}{\milli\second} per frame, and the maximum of four at 30 fps (120 frames per second) leaves \SI{8.33}{\milli\second}. These are averages a single processing resource has to hold rather than per-frame deadlines, and a tolerance to jitter in detection timing does not relax the average: a mean service time above \SI{25.0}{\milli\second} sustains fewer than 40 frames per second. Raising the per-frame time would require conceding dropped frames or adding a second processing resource, and we evaluate neither. What this paper measures, however, is the \emph{classification} that follows extraction, which is not tied to the frame period; its time constraint and the resulting throughput are treated in \S\ref{sec:optimal}.

\section{Resolution-native retraining and its architecture dependence}
\label{sec:adapt}

\begin{figure}[t]
\centering
\includegraphics[width=\textwidth]{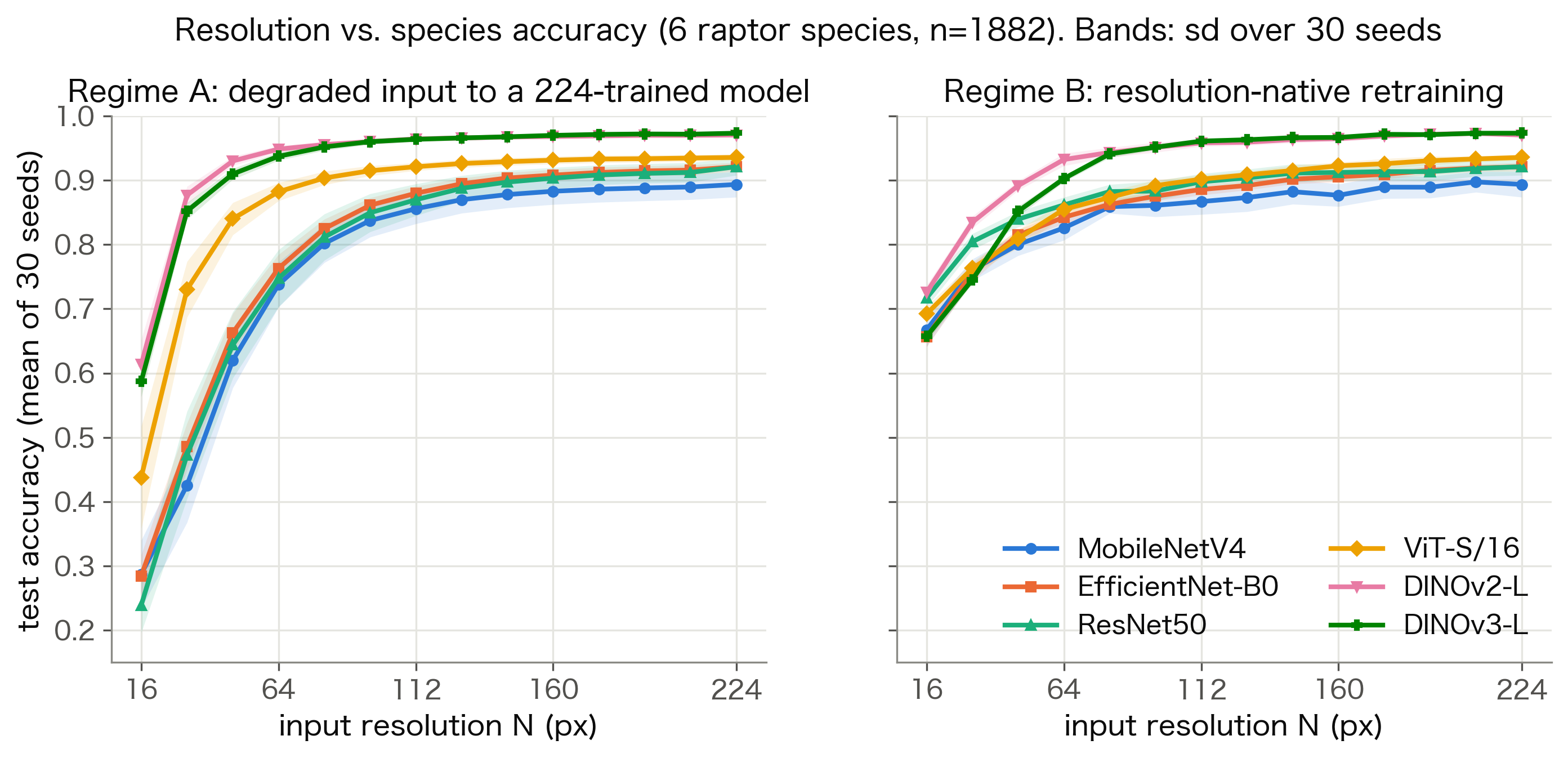}
\caption{Accuracy vs.\ resolution for the six architectures. Left: regime~A (degraded input to a 224-trained model). Right: regime~B (resolution-native retraining). Bands show the standard deviation over 30 seeds.}
\label{fig:acc}
\end{figure}

Figure~\ref{fig:acc} and Table~\ref{tab:adapt} give the gain of regime~B over regime~A. CNNs improve substantially when retrained at low resolution: ResNet50 goes from $0.2391$, closer to the $1/6$ chance level than to its $N{=}224$ accuracy, to $0.7169$ at $N{=}16$, a gain of $+47.8$ percentage points. \textbf{These results suggest that much of the low-resolution degradation observed for CNNs under regime~A is recovered by the regime~B pipeline, rather than reflecting an inherent inability to process small inputs.} Since the two regimes differ jointly in training resolution, weights and input geometry (\S\ref{sec:design}), the gain quantifies the distance between the two pipelines; it does not isolate the train--evaluation resolution mismatch as the single cause. The self-supervised ViT-L models move the other way: the mean difference is negative at most resolutions above $N{=}16$---at every tested level from $N{=}32$ to $N{=}160$ for both DINOv2-L and DINOv3-L---although small positive mean differences occur at some higher resolutions; ViT-S/16 is negative from $N{=}48$ to $N{=}208$ (at $N{=}224$ the two regimes coincide). One interpretation is that fine-tuning on low-resolution inputs degrades the pretrained representation, though we did not test this directly. At $N{=}16$, retraining helps every model ($+7.0$ to $+47.8$ percentage points); at that extreme the pretrained representation provides little benefit.

\begin{table}[t]
\centering
\caption{Gain of regime~B over regime~A (points, mean of 30 seeds).}
\label{tab:adapt}
\begin{tabular}{lrrrrr}
\toprule
Model & $N{=}16$ & 32 & 48 & 96 & 224 \\
\midrule
MobileNetV4 & \textbf{+38.0} & +33.5 & +18.0 & +2.4 & 0.0 \\
EfficientNet-B0 & \textbf{+37.2} & +27.1 & +15.2 & +1.4 & 0.0 \\
ResNet50 & \textbf{+47.8} & +33.2 & +19.6 & +3.4 & 0.0 \\
ViT-S/16 & +25.4 & +3.3 & $-$3.1 & $-$2.3 & 0.0 \\
DINOv2-L & +11.3 & $-$4.2 & $-$3.8 & $-$0.9 & 0.0 \\
DINOv3-L & +7.0 & \textbf{$-$10.7} & $-$5.8 & $-$0.9 & 0.0 \\
\bottomrule
\end{tabular}
\end{table}

Using many seeds matters here. The sign of the gain agrees across all 30 seeds for the three CNNs at $N{=}16$--$80$ and for the two self-supervised ViT-L models at $N{=}32$--$64$; at $N{=}80$ the ViT-L direction holds for 28 of 30 seeds for DINOv2-L and 29 of 30 for DINOv3-L, and from $N{=}96$ upward only 14--30 of 30 agree (over that range the mean differences themselves are at most 3.4 percentage points). Across all six architectures at once the sign of the mean difference agrees only at $N{=}16$; with three seeds it appears to agree over a much wider range. These counts describe the stability of the sign within the 30 observed seeds; they are not a test of whether the mean difference is zero, and a single seed cannot make the distinction at all.

How many seeds does it take for the sign to settle? We answered this over $k$-seed subsets of the 30 seeds: all 4{,}060 subsets for $k{=}3$ and all 142{,}506 for $k{=}5$ were enumerated exhaustively, while for $k{=}10$ and $k{=}20$, where the number of subsets exceeds 200{,}000, we sampled 200{,}000 with a fixed seed. The 78 pairs are the six architectures at 13 resolutions ($N{=}224$ is excluded because regimes~A and~B receive identical input there, so the difference is identically zero and carries no sign). Stratifying by effect size splits the picture. For the 41 pairs whose 30-seed mean differs by at least one percentage point, the \emph{median} agreement is already 100\% at $k{=}1$---though this is a median, not a statement about every pair: 26 of the 41 agree across all 30 seeds and the lowest agrees in 66.7\%---whereas for the 37 pairs below one point the median is 63.3\% at $k{=}1$, 77.6\% at $k{=}5$, 89.2\% at $k{=}10$ and 99.4\% at $k{=}20$. The risk of using few seeds is less that the sign comes out wrong than that the effect \emph{looks} consistent: at $k{=}3$, 50.3\% of subsets are unanimous (the median over pairs; the mean over pairs is 60.8\%), and 3.1\% of those unanimous subsets disagree with the 30-seed mean when the counts are pooled over all pairs---1.9\% if one instead reports the joint event ``unanimous and opposite'' as a fraction of all subsets, and 7.9\% on average, up to 60.6\%, if one averages the conditional fraction over pairs. Unanimity here means all $k$ differences strictly positive or all strictly negative; 34 of the 2{,}340 differences are exactly zero and are not counted as negative. Median, mean-over-pairs, joint and conditional are four different numbers, and at $k{=}20$ the unanimity rate is 0.0\% by median against 35.3\% by mean. The number of seeds required is thus not fixed but depends on the effect size one wants to claim. Two caveats: this is a description of sign stability, not a hypothesis test; and because the reference mean is computed from the same 30 seeds, it describes stability \emph{within the observed 30} and does not certify how many seeds an independent replication would need.

Sign stability is not the same as selection stability. Re-running the selection rule of \S\ref{sec:optimal} itself on $k$-seed validation means over the 29 seeds shared by all six models, using the 84 base candidates (exhaustive where $\binom{29}{k}\le2{,}000$, otherwise 2{,}000 sampled subsets) shows that the smallest $k$ at which the selection agrees with the common-29-seed reference 95\% of the time varies widely: 1 for the 0.95 target, 10 for 0.90, 26 for the \SI{33.3}{\milli\second} budget (for sampled $k$ these agreement rates are estimates under our draw of 2{,}000 subsets---$k{=}9$ at the 0.90 target already reaches 94.65\%---so a minimum $k$ near the 95\% boundary can shift with the draw). At the 0.93 target the reference itself differs from Table~\ref{tab:acctarget}: the validation mean of ResNet50 at $N{=}224$ is 0.9302 over 30 seeds but 0.9298 over the 29 seeds without seed 42, so the common-seed reference is DINOv2-L at $N{=}64$ at $3.1\times$ the classification time---a flip that itself shows how close the selection boundary sits to the target. Even the 29 leave-one-out subsets at $k{=}28$ agree with that reference in only 21 of 29 cases (72.4\%), the other eight all returning ResNet50 at $N{=}224$, and self-agreement at $k{=}29$ does not certify stability in an independent replication. On the extended 98-candidate set the reference instead becomes ViT-S/16 FP32 at $N{=}192$, but the agreement is again 21 of 29. The number of seeds required therefore depends not only on the effect size but also on the distance between the selection boundary and the target; the full grid is published as \texttt{selection\_stability.json}.

\begin{figure}[t]
\centering
\includegraphics[width=0.86\textwidth]{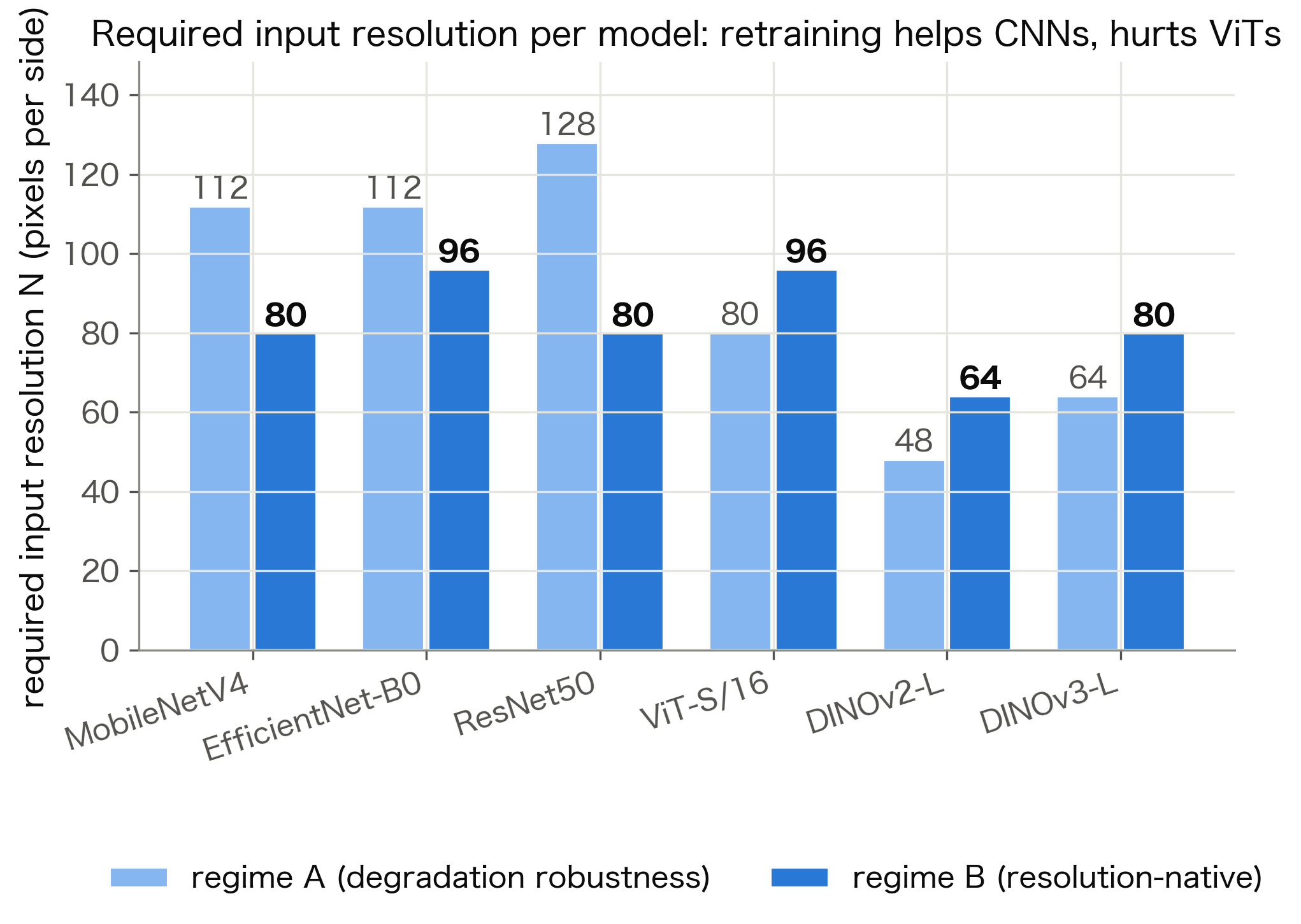}
\caption{Minimum required input side length for each model. Retraining lowers the requirement for CNNs and raises it for the ViT models.}
\label{fig:pixreq}
\end{figure}

We define the \emph{minimum required input side length} as the smallest $N$ whose accuracy stays within 5 percentage points of the same model's $N{=}224$ accuracy under the same regime (Figure~\ref{fig:pixreq}). Under regime~A the requirement is $N{=}48$--$64$ for the two self-supervised ViT-L models but $N{=}112$--$128$ for the CNNs, a two- to three-fold spread across families (DINOv2-L 48, DINOv3-L 64, MobileNetV4 and EfficientNet-B0 112, ResNet50 128, ViT-S/16 80). Regime~B lowers the CNN requirement to $N{=}80$--$96$ (MobileNetV4 and ResNet50 80, EfficientNet-B0 96) while raising it for the ViT models (DINOv2-L 64, DINOv3-L 80, ViT-S/16 96).

\section{Input resolution, processing time, and the role of the preprocessing path}
\label{sec:speed}

Table~\ref{tab:latency} shows FP16 inference time on Orin Nano. The total input pixel count differs by a factor of 196 between $N{=}16$ and $N{=}224$, yet latency changes by only $1.48$--$1.95\times$. If latency scaled linearly with total input pixel count, the ratio would be 196. GPU utilisation stays at 44--68\% even during inference, suggesting that kernel-launch and memory-transfer overheads contribute substantially.

\begin{table}[t]
\centering
\caption{Orin Nano FP16 inference time (\si{\milli\second}; median of 30 runs).}
\label{tab:latency}
\begin{tabular}{lrrrr}
\toprule
Model & $N{=}16$ & $N{=}112$ & $N{=}224$ & $r224/r16$ \\
\midrule
MobileNetV4 & 0.966 & 1.034 & 1.426 & \textbf{1.48} \\
EfficientNet-B0 & 1.284 & 1.427 & 2.113 & 1.65 \\
ResNet50 & 0.996 & 1.164 & 1.858 & 1.86 \\
ViT-S/16 & 1.211 & 1.310 & 2.359 & 1.95 \\
\bottomrule
\end{tabular}
\end{table}

What happens once preprocessing is included depends entirely on \emph{which} preprocessing is meant. Because this determines the paper's conclusion, we measured both paths on the same device (Orin CPU, $n=200$).

\paragraph{Offline evaluation path.} Here each individual is decoded from its own image file, as an evaluation script does. Decoding (\SI{2.67}{\milli\second}) and resizing (\SI{1.95}{\milli\second}) are then \emph{independent of resolution}, and this fixed \SI{4.6}{\milli\second} accounts, averaged over the 14 resolutions, for 80\% of the \SI{5.19}{}--\SI{6.80}{\milli\second} preprocessing time. Summed over preprocessing and GPU compute, MobileNetV4 costs \SI{7.40}{\milli\second} at $N{=}224$ (5.98 preprocessing + 1.43 GPU) and \SI{6.41}{\milli\second} at $N{=}80$ (5.43 + 0.98): reducing the total input pixel count by a factor of 7.84 saves only 13.5\%.

\paragraph{Is the summed time the real time?} The classification time we report is preprocessing plus GPU compute, and for the ViT-L-scale models the GPU term is a sum over the parts of a split chain, which omits host-side dispatch, synchronisation and the handover between parts. We therefore measured the \emph{wall-clock of a single classification} (host-to-device copy, all parts, device-to-host copy and synchronisation---CPU preprocessing is not inside this interval) for five configurations over $n=1{,}882$ images $\times$ three repetitions, discarding the first 200 of each repetition as warm-up, which leaves $1{,}682\times3$ timings per configuration. The median exceeds the summed model by \SI{0.22}{}--\SI{0.31}{\milli\second}, and by the same amount for a single-engine ResNet50 as for a five-part DINOv2-L chain (\SI{+0.223}{} at $N{=}112$ and \SI{+0.258}{\milli\second} at $N{=}224$ against \SI{+0.224}{} for DINOv2-L at $N{=}64$ and \SI{+0.314}{\milli\second} at $N{=}224$); DINOv3-L at $N{=}224$ is \SI{1.44}{\milli\second} \emph{faster} than the sum. The overhead is the same for one part as for five, so we observed no clear growth with the number of parts---though a single-versus-multi-part comparison alone cannot separate dispatch and copy costs from a cost of splitting. For the configuration selected under the \SI{10}{\milli\second} budget, DINOv2-L at $N{=}64$, this interval is \SI{9.28}{\milli\second} at the median and \SI{9.44}{\milli\second} at the 95th percentile; adding the separately measured preprocessing figure of \SI{0.35}{\milli\second} gives \SI{9.63}{} and \SI{9.79}{\milli\second}, still inside the budget. These two figures are therefore estimates, not percentiles of a continuously measured total that would include the variability of preprocessing itself.

\paragraph{Deployment path.} A deployed system does not work this way. The 4K frame has already been decoded once by the detector, which runs on every frame; the classifier receives an in-memory array and only crops the bounding box, resizes it to $N\times N$ and normalises it. No per-individual decode occurs. Measured this way, per-individual preprocessing is \SI{0.136}{}--\SI{1.774}{\milli\second}---between $1/41$ and $1/5$ of the offline path---and, because the dominant terms are the $O(N^2)$ normalisation and resize, it \emph{does} depend strongly on $N$. The same MobileNetV4 costs \SI{2.65}{\milli\second} at $N{=}224$ and \SI{1.42}{\milli\second} at $N{=}80$: a saving of 46.2\%, not 13.5\% (Table~\ref{tab:prep}); the other three models save 47.2--53.0\%.

\begin{table}[htbp]
\centering
\caption{Effect of the assumed preprocessing path (Orin Nano, $n=200$). The classification columns are estimated preprocessing plus GPU compute for one individual; GPU time is identical in both rows, so only the preprocessing model differs. The last column is the saving from $N{=}224$ to $N{=}80$ within each path, not a per-resolution quantity.}
\label{tab:prep}
\begin{tabular}{lrrrrr}
\toprule
& \multicolumn{2}{c}{Preprocessing [ms]} & \multicolumn{2}{c}{MobileNetV4 classification [ms]} & Saving \\
\cmidrule(lr){2-3}\cmidrule(lr){4-5}\cmidrule(lr){6-6}
Path & $N{=}224$ & $N{=}80$ & $N{=}224$ & $N{=}80$ & $224\rightarrow80$ \\
\midrule
Offline evaluation & 5.98 & 5.43 & 7.40 & 6.41 & 13.5\% \\
Deployment & \textbf{1.22} & \textbf{0.45} & \textbf{2.65} & \textbf{1.42} & \textbf{46.2\%} \\
\bottomrule
\end{tabular}
\end{table}

What a deployment is constrained by is the second row. Reducing input resolution therefore does buy time at the edge. The commonly reported observation that ``preprocessing sets a floor, so lowering resolution does not help'' embeds an overhead specific to file-based evaluation and cannot be used to size a deployed system.

\subsection{Latency ratios including the ViT-L-scale models}

\begin{figure}[t]
\centering
\includegraphics[width=\textwidth]{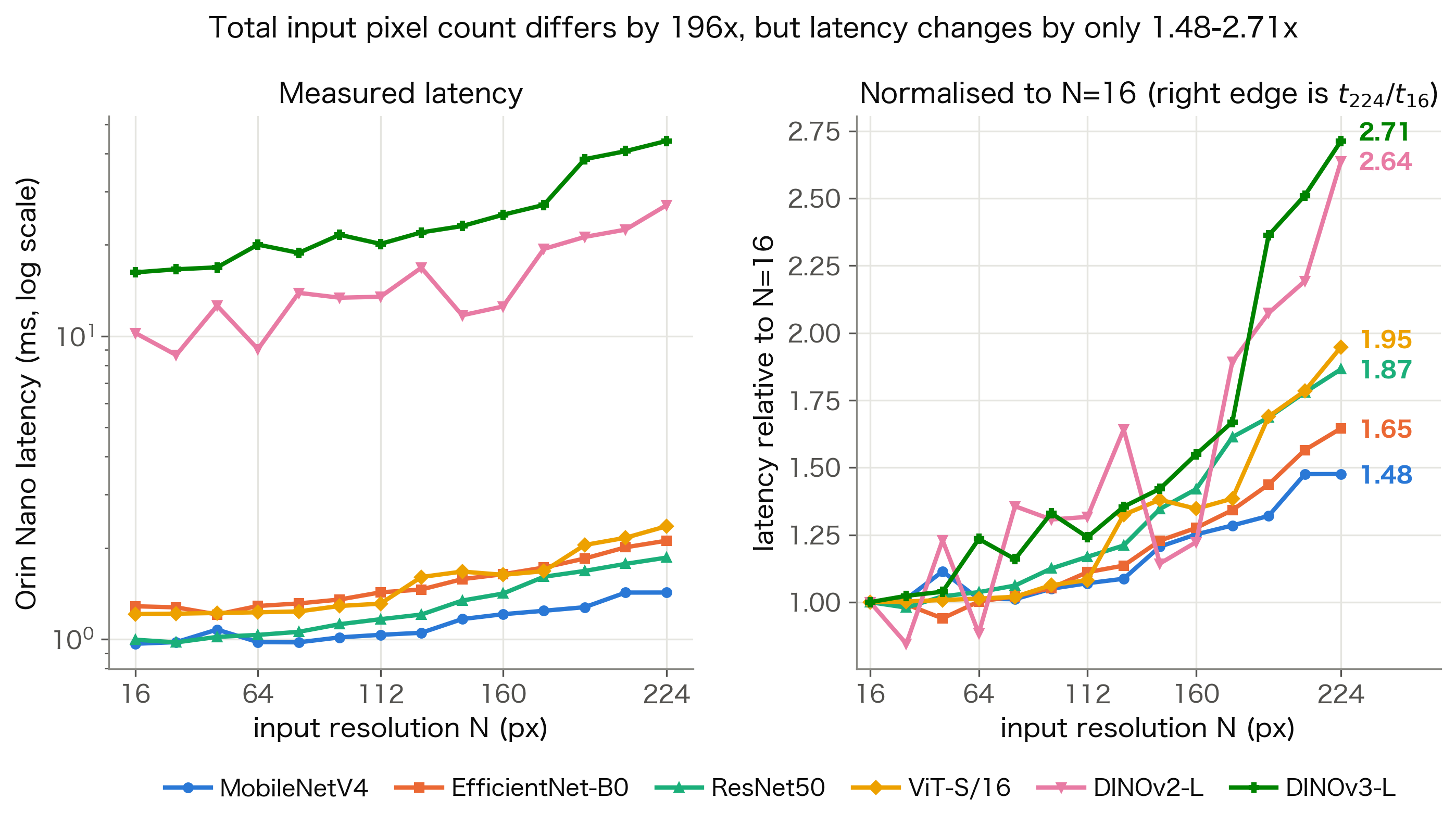}
\caption{Latency against input resolution for all six models. Left: measured latency. Right: normalised to $N{=}16$, so the right-hand edge is $t_{224}/t_{16}$.}
\label{fig:scale}
\end{figure}

We measured both ViT-L models at all 14 resolutions (Table~\ref{tab:vitlres}, Figure~\ref{fig:scale}). The H200 numbers rest on three runs and vary more than the Orin ones, so we treat them as indicative and do not draw architecture-level conclusions from the edge/server ratio. Defining the latency ratio as $t_{224}/t_{16}$, \emph{all six models fall in the range $1.48$--$2.71$}, including the two ViT-L-scale models (DINOv2-L $2.64$, DINOv3-L $2.71$). Against a 196-fold difference in total input pixel count this is two orders of magnitude smaller, so \textbf{inference time is not proportional to the total input pixel count}. The same DINOv2-L on an H200 shows $1.45\times$; the same tendency is therefore observed on the server side. The ratio does depend on the precision configuration: rebuilding DINOv2-L entirely in FP32 (\S\ref{sec:vitl}) raises $t_{224}/t_{16}$ to $4.18$, consistent with the reading that the more of a network runs in half precision, the larger the resolution-independent fixed cost becomes relative to the total.

\begin{table}[htbp]
\centering
\caption{ViT-L on the edge at all 14 resolutions. $N'$ is the side length actually fed to the model: DINOv2-L is patch-14, so the requested $N\in\{16,32,\dots,224\}$ is rounded to a multiple of 14, whereas DINOv3-L is patch-16 and $N'=N$. Rows of the two blocks correspond to the same requested $N$. Accuracy is top-1 measured on the device; agreement is argmax agreement with server FP32. Every configuration was run on all $n=1{,}882$ test images and verified to depend on its input. The accuracy and agreement columns come from a \emph{single checkpoint} (seed 42), not from a multi-seed mean; the 29-seed means are given in \S\ref{sec:agreement}.}
\label{tab:vitlres}
\begin{tabular}{lrrrrr}
\toprule
$N'$ & Orin (\si{\milli\second}) & H200 (\si{\milli\second}) & edge/server & accuracy & agreement \\
\midrule
\multicolumn{6}{l}{\emph{DINOv2-L, 5-way split, FP32 only on the last part; $t_{224}/t_{16}=2.64$}} \\
14 & \textbf{10.240} & 1.099 & \textbf{9.3} & 0.7343 & 99.79\% \\
28 & 8.657 & 1.100 & 7.9 & 0.8443 & 99.79\% \\
42 & 12.596 & 1.075 & 11.7 & 0.8895 & 97.93\% \\
70 & 9.056 & 1.076 & 8.4 & 0.9304 & 98.94\% \\
84 & 13.889 & 1.167 & 11.9 & 0.9501 & 99.15\% \\
98 & 13.384 & 1.115 & 12.0 & 0.9469 & 99.04\% \\
112 & 13.482 & 1.294 & 10.4 & 0.9522 & 99.42\% \\
126 & 16.810 & 1.230 & 13.7 & 0.9501 & 99.10\% \\
140 & 11.710 & 1.273 & 9.2 & 0.9676 & 99.68\% \\
154 & 12.523 & 1.228 & 10.2 & 0.9729 & 99.73\% \\
182 & 19.387 & 1.440 & 13.5 & 0.9617 & 99.31\% \\
196 & 21.236 & 1.391 & 15.3 & 0.9718 & 99.63\% \\
210 & 22.451 & 1.408 & 15.9 & 0.9665 & 99.26\% \\
224 & \textbf{27.014} & 1.597 & \textbf{16.9} & 0.9724 & 99.79\% \\
\midrule
\multicolumn{6}{l}{\emph{DINOv3-L, 4-way split, all parts FP32; $t_{224}/t_{16}=2.71$}} \\
16 & \textbf{16.228} & 1.093 & \textbf{14.8} & 0.6594 & 99.95\% \\
32 & 16.608 & 1.076 & 15.4 & 0.7476 & 99.95\% \\
48 & 16.855 & 1.079 & 15.6 & 0.8613 & 99.89\% \\
64 & 20.036 & 1.108 & 18.1 & 0.9075 & 100.00\% \\
80 & 18.819 & 1.085 & 17.3 & 0.9346 & 100.00\% \\
96 & 21.589 & 1.162 & 18.6 & 0.9538 & 99.95\% \\
112 & 20.144 & 1.160 & 17.4 & 0.9660 & 100.00\% \\
128 & 21.980 & 1.290 & 17.0 & 0.9633 & 99.95\% \\
144 & 23.063 & 1.309 & 17.6 & 0.9692 & 100.00\% \\
160 & 25.139 & 1.322 & 19.0 & 0.9697 & 100.00\% \\
176 & 27.083 & 1.310 & 20.7 & 0.9740 & 100.00\% \\
192 & 38.337 & 1.478 & 25.9 & 0.9718 & 99.95\% \\
208 & 40.741 & 1.470 & 27.7 & 0.9740 & 100.00\% \\
224 & \textbf{44.028} & 1.456 & \textbf{30.2} & 0.9729 & 100.00\% \\
\bottomrule
\end{tabular}
\end{table}

On-device accuracy stays within $-0.16$ to $+0.59$ percentage points of server FP32, with no systematic loss (this table is a single seed, 42; the 29-seed means are given in \S\ref{sec:agreement}). Agreement is uniformly high across all 14 resolutions---97.9--99.8\% for DINOv2-L and 99.9--100\% for DINOv3-L---with no monotone trend in resolution: these ViT-L configurations reproduce the server's decisions almost exactly on the edge device.

\begin{figure}[t]
\centering
\includegraphics[width=\textwidth]{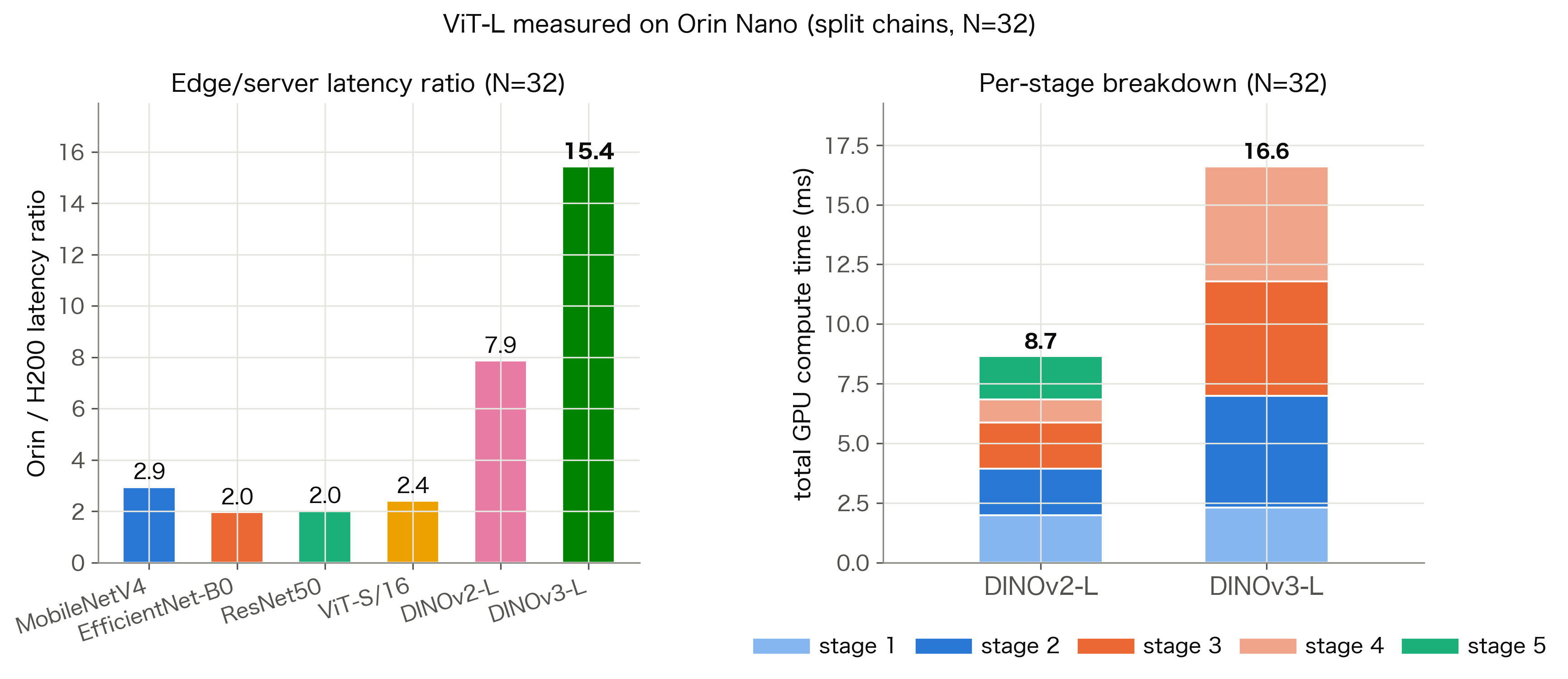}
\caption{Left: edge/server latency ratio, all at $N{=}32$. Right: per-stage breakdown of the deployed split chains at $N{=}32$.}
\label{fig:edge}
\end{figure}

The edge/server ratio itself moves by a factor of 3.8 across resolutions ($7.9 \rightarrow 30.2$). Even at a fixed $N{=}224$ it spans $2.9\times$ (EfficientNet-B0) to $30.2\times$ (DINOv3-L), a ten-fold spread, whereas latencies across the evaluated models differ by only $3.3\times$ on the H200 (Figure~\ref{fig:edge}, left). Server benchmarks do not extrapolate to the edge.

\section{Selecting the input resolution from accuracy and processing time}
\label{sec:optimal}

Combining the accuracy of \S\ref{sec:adapt} with the measured time of \S\ref{sec:speed} answers the question posed in the introduction: which input resolution $N$ delivers high accuracy at the shortest processing time, and what are that accuracy and time? Throughout this section, \emph{classification time} denotes the time of one classification on the deployment path, since that is what a deployment is actually constrained by (\S\ref{sec:speed}). It is an \emph{estimate} assembled from two separately measured terms---CPU preprocessing for one individual and GPU compute---and it excludes detection, tracking and frame decoding, so it is not a continuously measured time for the whole system; the wall-clock check of \S\ref{sec:speed} bounds the gap for the classification stage itself. Offline-path figures are given in parentheses for reference.

\subsection{Selection by target accuracy}

For each accuracy target, Table~\ref{tab:acctarget} lists the configuration that meets it in the shortest estimated classification time. Which configuration wins depends on the target, and it turns on the choice of model far more than on the choice of $N$.

\begin{table}[htbp]
\centering
\caption{Fastest configuration meeting each accuracy target, among the 84 base candidates. The time column is the \emph{estimated classification time} (preprocessing plus GPU compute for one individual) on the deployment path, with the offline-evaluation figure in parentheses; it excludes detection, tracking and frame decoding and is not a measured whole-system time. $N'$ is the side length actually fed to the model; DINOv2-L rounds to a multiple of 14. Accuracy is measured on the deployed Orin engine (30 seeds; 29 for the ViT-L-scale models). Configurations are selected on the validation accuracy shown and the test accuracy is reported alongside; the two rows marked $\dagger$ meet the target on validation but fall slightly short on test.}
\label{tab:acctarget}
\begin{tabular}{llrrrrr}
\toprule
Target & Model & $N$ & $N'$ & Classification [ms] & Validation & Test accuracy \\
\midrule
0.80 & ResNet50 & 32 & 32 & \textbf{1.16} (6.17) & 0.8134 & 0.8049 \\
0.85 & ResNet50 & 64 & 64 & 1.38 (6.38) & 0.8694 & 0.8614 \\
0.90 & ResNet50 & 112 & 112 & 1.85 (6.84) & 0.9051 & 0.8980$^\dagger$ \\
0.93 & ResNet50 & 224 & 224 & 3.08 (7.84) & 0.9302 & 0.9213$^\dagger$ \\
0.95 & DINOv2-L & 144 & 140 & 12.70 (17.68) & 0.9653 & 0.9630 \\
0.96 & DINOv2-L & 144 & 140 & 12.70 (17.68) & 0.9653 & 0.9630 \\
\bottomrule
\end{tabular}
\end{table}

The target most often required in practice, 0.90, is met on validation at exactly $N{=}112$ (ResNet50, \SI{1.85}{\milli\second}, test accuracy 0.8980); only when 0.95 or more is demanded does the selection move to DINOv2-L at $N{=}144$ (\SI{12.70}{\milli\second}), while 0.80--0.85 are already met at $N{=}32$--$64$. Raising $N$ becomes necessary only when the target is pushed to 0.93 or above. These selections are made among the 84 base candidates and on the deployment path; the extended candidate set discussed below moves the recommendation over part of that range.

The accuracies in Table~\ref{tab:acctarget} are those of the engine that is actually deployed. To ask how much this matters for \emph{selection}, we held the selection split at validation and changed only the accuracy used to select---server FP32 or deployed engine, with the same seed set on both sides for each model (the common 29 seeds for the ViT-L-scale models, 30 otherwise)---sweeping the target from 0.800 to 0.975 in steps of 0.001. The selected configuration differs at 26 of the 176 targets (14.8\%): four isolated targets (0.814, 0.870, 0.887 and 0.890, each moving one resolution step within ResNet50) and the ranges 0.919--0.927, 0.931--0.938, 0.966--0.968, 0.972 and 0.975; the model family changes at 0.920--0.927, 0.931--0.938, 0.972 and 0.975. The largest is at 0.931--0.938, where selecting on server FP32 returns ViT-S/16 at $N{=}224$ (validation 0.9389, \SI{3.58}{\milli\second}) whose deployed test accuracy is only 0.8801, while selecting on the deployed engine returns DINOv2-L at $N{=}64$ for 0.931--0.932 and at $N{=}144$ for 0.933--0.938---a different model family and $2.6\times$ to $3.5\times$ the time. At 0.966--0.968, 0.972 and 0.975 the deployed validation of DINOv2-L sits 0.02--0.24 points below server FP32 over the same 29 seeds, which moves the choice up within DINOv2-L or across to DINOv3-L. At 0.90 and 0.93, by contrast, both criteria return the same configuration (ResNet50 at $N{=}112$ and $N{=}224$): ViT-S/16 at $N{=}112$ does reach 0.9002 on server validation but takes \SI{1.996}{\milli\second} against \SI{1.850}{\milli\second} for ResNet50, so it is never the fastest. A table that pairs server-side accuracy with edge-side latency can describe a configuration that does not exist, but whether that reaches the selection depends on the target. One dependency remains: the ViT-S/16 entries here are its FP16 engine. We therefore rebuilt ViT-S/16 as an all-FP32 engine and measured all 14 resolutions over 30 seeds on the device: deployed validation tracks server FP32 within $-0.07$ to $-0.01$ points, at $2.5\times$ the FP16 GPU time (\SI{5.92}{} versus \SI{2.36}{\milli\second} at $N{=}224$). Admitting these 14 as candidates---an extended set of 98 against the 84 used everywhere else---replaces the recommendation over the whole 0.931--0.938 band---DINOv2-L at $N{=}64$ and $N{=}144$---with ViT-S/16 FP32 at $N{=}192$--$224$, whose classification time is 53--71\% of the former, and moves the \SI{10}{\milli\second}-budget recommendation to ViT-S/16 FP32 at $N{=}224$ (validation 0.9386, \SI{7.14}{\milli\second}). The Pareto sets grow from 25 to 27 on GPU compute and from 23 to 25 on the deployment path (both admitting $N{=}192$, 208 and 224) and from 20 to 21 offline (admitting $N{=}192$ and 224 only: at \SI{12.03}{\milli\second} $N{=}208$ is dominated by $N{=}224$ at \SI{11.90}{\milli\second}), and in every case DINOv2-L at $N{=}64$ is the only configuration to leave. The offline-path \SI{10}{\milli\second} choice remains ResNet50 at $N{=}224$, so the path dependence of \S\ref{sec:speed} persists. The recommendations are thus conditional on the precision assignment given to each model---measurably so.

\paragraph{Class imbalance and per-class performance.} Accuracy can hide misses on rare classes, so we also computed per-class metrics for the selected configurations, averaging the metric over seeds rather than pooling the predictions. The imbalance is mild---the largest class has 2.54 times the support of the smallest---and macro-F1 comes out slightly \emph{above} accuracy (0.9040 against 0.8980 for ResNet50 at $N{=}112$, 0.9652 against 0.9630 for DINOv2-L at $N{=}144$), so accuracy is not concealing a minority-class failure here. The hardest class is the white-tailed eagle---recall 0.859 for ResNet50 at $N{=}112$, rising to 0.947 for DINOv2-L at $N{=}144$---and its confusions are consistently with the golden eagle, the other large dark eagle in the set. ViT-S/16 at $N{=}224$ behaves differently, with the golden eagle worst at 0.789 and confused with the northern goshawk.

\paragraph{Uncertainty on the central configurations.} Two sources of variation should not be mixed. \emph{Retraining} variability, the standard deviation over the 30 seeds, is 1.25 points for ResNet50 at $N{=}112$, 0.51 for DINOv2-L at $N{=}144$ and 3.80 for ViT-S/16 at $N{=}224$---the last being about three times the ResNet50 figure and seven and a half times the DINOv2-L one, and a reason to distrust single-seed comparisons involving that model. \emph{Sampling another set of birds} is a different question, and one that must be answered at the level of \emph{groups} rather than images, because crops of one photograph and adjacent frames of one video are not independent. Resampling the 1{,}720 groups of the test set with replacement (2{,}000 replicates, seed-averaged per-image correctness; the grouping rule includes the video identifier---under the coarser rule that misses it the count is 1{,}834 and the interval widths differ by less than 0.08 points) gives 95\% intervals of $[0.8870, 0.9084]$ for ResNet50 at $N{=}112$, $[0.9118, 0.9305]$ at $N{=}224$, $[0.9217, 0.9393]$ for DINOv2-L at $N{=}64$ and $[0.9563, 0.9696]$ at $N{=}144$. The $1{,}882 \times 30$ predictions are not 56{,}460 independent observations. Applying the same resampling to the paired differences, over the seeds common to both configurations (29 seeds, 43--71, for the model change and the 10 ms pair; 30 for the same-model resolution change), gives $+5.97$ points $[+5.11, +6.80]$ for changing the model at $N{=}112$, $+2.33$ $[+1.83, +2.87]$ for raising $N$ from 112 to 224, and $+0.98$ $[+0.29, +1.68]$ between the two configurations selected under the \SI{10}{\milli\second} budget; none of the three intervals contains zero, though the last is narrow. These paired means differ by construction from the differences of the per-configuration means quoted elsewhere (5.93 and 0.94 points).

Configurations are selected on validation and reported on test. Because the two splits are kept separate, the rows for 0.90 and 0.93 meet the target on validation but fall slightly short on test (0.9051 versus 0.8980, and 0.9302 versus 0.9213). Such a shortfall can occur when selection and reporting use different splits: re-selecting on test so that the target is met there would introduce selection bias, and we prefer the former.

What separates the targets is mainly the model, not the resolution. Two ways of spending time are available at that point, and they are not equally productive. Replacing ResNet50 with DINOv2-L, at the same $N{=}112$, raises accuracy from 0.8980 to 0.9573 ($+5.93$ points, both measured on the deployed engine); keeping ResNet50 and raising $N$ from 112 to 224---a fourfold increase in total pixel count---gives only $+2.33$ points. In absolute accuracy, then, changing the model is the larger step. The costs run the other way, however: changing the model takes the estimated classification time from 1.85 to \SI{14.17}{\milli\second} ($7.7\times$), whereas raising the resolution takes it only from 1.85 to \SI{3.08}{\milli\second} ($1.66\times$). The two levers are therefore compared at very different costs, and the larger accuracy increment does not by itself make the model change the better use of a fixed time budget; which one wins under a budget is what the selection of \S\ref{sec:budget} decides.

\subsection{Selection under a time constraint}
\label{sec:budget}

The same data can be read from the other side, as a limit on the time allowed for a \emph{single classification}. As noted in \S\ref{sec:design}, classification is not tied to the frame period; neither is it unlimited, since it shares the machine with detection. We therefore consider constraints from \SIrange{10}{100}{\milli\second} as a guide. Allowing \SI{33.3}{\milli\second}, the most accurate configuration is DINOv3-L at $N{=}176$ (0.9715, \SI{28.45}{\milli\second}); tightening the constraint to \SI{10}{\milli\second} leaves DINOv2-L at $N{=}64$ (0.9307, \SI{9.40}{\milli\second}), 4.1 points lower.

Tightening further still leaves choices. The fastest configuration is MobileNetV4 at $N{=}16$ at \SI{1.10}{\milli\second}, and the time decreases as $N$ decreases. On the offline evaluation path no configuration was faster than \SI{6.17}{\milli\second}, which substantially narrows the set admitted under a \SI{10}{\milli\second} constraint. The best configuration under a tight time constraint therefore depends on the assumed preprocessing path.

If instead each detected individual is classified once, the time is not tied to the frame period: \SI{50}{\milli\second} admits DINOv3-L at $N{=}208$ (0.9733, \SI{42.52}{\milli\second}), though it is only 0.01 points above DINOv2-L at $N{=}208$ while taking $1.8\times$ the time.

\subsection{Number of individuals processable within one camera's field of view}
\label{sec:throughput}

Classification runs only on extracted candidates, so it is not tied to the frame period. It is, however, bounded: within a unit, the classifications for all four cameras share one Orin Nano. Writing $T$ for the estimated classification time of one individual in milliseconds and $c$ for the number of classifications per individual (normally one, at most four including the escalations), one camera can process $1000/(4cT)$ individuals per second. This is the service rate of the classifier alone: it assumes the classifier has the accelerator to itself, and does not reserve a share for the detector, which runs on the same device. With a fraction $\rho$ of the device available to classification the rate scales by $\rho$.

At $N{=}112$ with $c{=}4$ the rates, in individuals per second, are 36.3 for MobileNetV4, 33.8 for ResNet50, 31.3 for ViT-S/16, 29.6 for EfficientNet-B0, 4.4 for DINOv2-L and 3.0 for DINOv3-L; with $c{=}1$ all of these are four times larger. In the two-unit configuration each unit serves its own four cameras, so the per-camera figure is unchanged. Multiplying by the dwell time in the field of view gives an \emph{average} number of individuals that can be handled under a steady arrival assumption; it is not a guarantee against loss when arrivals cluster, and it ignores queueing and re-classification of tracked individuals. ResNet50, the configuration selected for the 0.90 target, and the ViT-L-scale models thus differ by a factor of 7.7 in the number of individuals they can handle, which corresponds to the accuracy difference between 0.90 and 0.96 (taking the lightest model, MobileNetV4, as the reference gives $8.2\times$). When many birds arrive at once, that gap may increase classification backlog under bursty arrivals; because we did not measure the arrival process, waiting times or the interference between classification and detection, we cannot convert it into a rate of missed extractions. Assuming the offline evaluation path compresses this ratio to 2.8, but only because a common overhead depresses both figures equally, so it understates the difference relevant to deployment.

\subsection{Constraints imposed on the optics by the input resolution}
\label{sec:optics}

The value of being able to lower $N$ is not confined to processing time. The side length an object occupies in the image is $N=f_{\mathrm{px}}S/D$, where $f_{\mathrm{px}}$ is the focal length in pixels, $S$ the physical size of the target and $D$ the range, so for a given camera the achievable range is inversely proportional to $N$. We take $S=\SI{1.5}{\metre}$, the wingspan of a black kite or white-tailed eagle.

\paragraph{Atmospheric limits on the attainable resolution (a design consideration, not a measurement).} It is worth separating three quantities that are easy to conflate: $N_{\mathrm{net}}$, the side length fed to the network; $n_{\mathrm{obj}}$, the number of sensor pixels the target subtends; and $M$, the number of \emph{independently resolvable} elements the optical path delivers. Resizing a blurred image to $224$ pixels is always possible, so a small $M$ does not forbid $N_{\mathrm{net}}{=}224$; what it limits is the information that reaches the network. Under turbulence the achievable angular resolution is limited to $\lambda/r_0$, where $r_0=(0.423k^2C_n^2L)^{-3/5}$ is the Fried parameter with $k=2\pi/\lambda$~\cite{fried}, and no increase in aperture improves it. We take the resolution element to be $\theta=\lambda/r_0$; that is, the coefficient of the Rayleigh-type criterion is set to unity here rather than to the factor 1.22 of the diffraction-limited circular-aperture form, which would reduce $M$ in proportion. Assuming $\lambda=\SI{550}{\nano\metre}$, a uniform $C_n^2=10^{-14}\,\mathrm{m}^{-2/3}$ over the path, the long-exposure average of~\cite{fried} and that Rayleigh-type criterion, $M$ at \SI{1}{\kilo\metre} works out to about 62 elements, and to about 15.5 under strong turbulence ($10^{-13}\,\mathrm{m}^{-2/3}$). These are order-of-magnitude figures that follow from the stated assumptions, not measurements of our system; we have not validated them against imagery taken at range. Read that way they say that a design premised on \emph{224 independent elements} at \SI{1}{\kilo\metre} is optimistic, and that studying the low-$N$ end of the sweep is worthwhile---not that identification at \SI{1}{\kilo\metre} is guaranteed, nor that any particular accuracy is attained there.

This ceiling constrains species identification from the image, not detection and tracking. Detection operates on background subtraction and succeeds on targets a few pixels across, so tracking can continue at ranges where image-based identification does not hold. Whether the shape of the flight trajectory and the speed distribution then support even a coarse species estimate is a separate question, which this study did not evaluate. The ceiling on $N$ therefore bounds the range over which image-based identification is informative while leaving detection and tracking available further out; we do not claim that it is immaterial to the mitigation decision.

\paragraph{Translating the requirement into a camera specification.} A requirement on $n_{\mathrm{obj}}$ can be converted into a lens specification, but not by comparing zoom ratios: an optical zoom ratio is defined relative to that camera's own wide end, so ratios from different cameras are not commensurable. The comparison has to be made on the telephoto-end horizontal field of view, focal length, sensor size and output resolution. For a \SI{1.5}{\metre} target at \SI{500}{\metre} and a camera delivering $3840$ pixels across a \SI{2.3}{\degree} telephoto field of view---a value we \emph{assume} as representative of the telephoto end of a 4K PTZ camera, rather than quote from a named product---the purely geometric projection is $n_{\mathrm{obj}} = 3840\times1.5/(2\times500\tan(2.3^{\circ}/2)) \approx 287$ pixels. A camera with these assumed characteristics would therefore project more than 224 pixels onto the target; whether a particular camera meets these specifications requires product-level verification at the design stage. This geometric figure ignores atmosphere, aberrations, target pose and motion blur, so it does not imply that 224 \emph{independent} elements, or any particular identification accuracy, are obtained. For reference, the diffraction-limited aperture at \SI{1}{\kilo\metre} is \SI{50}{\milli\metre} for 112 elements and \SI{100}{\milli\metre} for 224.

\subsection{Pareto-optimal configurations}

Across the 84 base configurations (six models $\times$ 14 resolutions, each with its default precision assignment; the extended 98-candidate set of \S\ref{sec:optimal} is used only there), 25 lie on the Pareto front when time is measured as GPU compute and 23 when measured as classification time on the deployment path; MobileNetV4 at $N{=}16$ is included in both. On the offline evaluation path only 20 remain, and the excluded configurations are the low-resolution ones. The cause is \emph{not} the resolution-independent overhead itself: adding the same constant to every candidate leaves the dominance relation unchanged, as we confirmed by adding \SI{3.0}{} and \SI{5.6}{\milli\second} uniformly to the GPU times and recovering exactly the same front. What a fixed cost changes is the percentage saving and the set of candidates that fit a budget. The front changes because preprocessing on this path is \emph{not monotone} in resolution---for MobileNetV4 it costs \SI{5.576}{\milli\second} at $N{=}16$ against \SI{5.191}{\milli\second} at $N{=}32$, so the more accurate $N{=}32$ dominates $N{=}16$---a residual that reflects implementation branches and measurement variation and that does not arise on the deployment path, where preprocessing follows $O(N^2)$. The margin is thin: the three excluded configurations trail their nearest dominating configurations by only 0.02--0.13 ms, whereas the per-image spread of preprocessing on this path is large (decode: median 2.1, p95 \SI{8.8}{\milli\second}, $n{=}200$) and the median decode time itself---nominally resolution-independent---measures \SI{2.39}{\milli\second} at $N{=}16$ against \SI{2.09}{\milli\second} at $N{=}32$; however, only summary statistics of the 200 repetitions are stored, and they do not allow us to assess the uncertainty of the mean preprocessing times, and hence of the front membership of these three boundary configurations. The 20-versus-23 set sizes are point estimates from this measurement whose reproducibility has not been verified. An incorrect assumption about the preprocessing path thus undervalues low-resolution configurations through a residual specific to that path (Fig.~\ref{fig:pareto}).

These results are summarised as follows. Among the 84 base candidates and at the 0.90 target, $N\approx112$ is the practical operating point for accuracy against time: raising $N$ beyond it yields little accuracy while increasing processing time. The decision shifts from the resolution to the model when the target exceeds 0.93, at which point a ViT-L-scale backbone becomes necessary \emph{within that candidate set}; on the extended 98-candidate set the 0.931--0.938 band is served instead by a ViT-S/16 FP32 engine (\S\ref{sec:optimal}), so this is a statement about the candidates on offer rather than about ViT-L-scale models in general. Lowering $N$ below this point, in contrast, yields two benefits---a reduction in latency (46.2\% from $N{=}224$ to $N{=}80$ on the deployment path) and relaxed optical requirements (\S\ref{sec:optics})---so $N$ may be reduced as far as the accuracy requirement permits. Whether such a model runs on this device is examined in \S\ref{sec:vitl}.

\section{Edge implementation of ViT-L-scale models}
\label{sec:vitl}

\subsection{Graph splitting and the granularity of precision control}

A 300M-parameter ViT-L fits this device without splitting. Built as a single unsplit engine at $N{=}32$, DINOv2-L yields a \SI{1213}{\mega\byte} engine whose argmax matches server FP32 on every one of the $n=1{,}882$ test images; DINOv3-L likewise builds as a single engine and agrees on 99.95\%. Splitting is therefore not a means of making the model fit.

Its significance lies in the granularity of precision control. As \S\ref{sec:const} shows, ViT-L activations exceed the FP16 range, and with the export and build procedure used here a single fused build has to run \emph{the whole graph} in FP32 to remain correct, including the intervals that could have run in half precision. That is an observation about these builds: we did not test whether finer-grained precision control inside an unsplit engine---per-channel scaling, layer-level precision hints or a different builder version---would avoid it. Built in FP16, the unsplit DINOv2-L returns output that does not depend on its input: 1{,}881 of 1{,}882 images share identical logits and accuracy collapses to 0.1201. Splitting at transformer-block boundaries confines FP32 to the intervals that need it and leaves the rest in FP16, which makes the deployed chain $1.85\times$ faster than the single-engine build (\SI{8.657}{} versus \SI{16.023}{\milli\second} at $N{=}32$). DINOv3-L requires FP32 throughout, so its margin is only $1.07\times$ (\SI{16.608}{} versus \SI{17.701}{\milli\second})---the speed benefit of splitting is large only where mixed precision is possible.

Splitting is feasible because only one tensor crosses each transformer-block boundary. TensorRT fuses the 2{,}430 ONNX operators of the transformer into a single \emph{foreign node}---a compiled unit the runtime treats as one opaque layer---so cutting at these boundaries reduces the fusion unit and makes per-interval precision assignment possible.

DINOv3-L needs one further step, because its graph cannot be parsed as exported. The exported ONNX contains two \texttt{If} operators---conditional branches---whose predicate is computed from the element count of an input tensor (\texttt{Greater(Sub(Size,$c_1$),$c_2$)}) and is therefore \emph{uniquely determined once the input shape is fixed}. The graph nonetheless carries them as run-time branches whose then- and else-branches yield outputs of different shapes, \texttt{[2]} versus \texttt{[1]}. Passing the graph through onnxruntime's basic optimisation before parsing folds the predicate under the fixed input shape. This takes \texttt{If} from two to zero and the node count from 3{,}600 to 2{,}714, leaving a graph of standard operators. All DINOv3-L numbers reported here are measured on graphs prepared this way.

\subsection{Breakdown of half-precision arithmetic caused by outlier activations}
\label{sec:const}

Splitting alone is not sufficient. \textbf{ViT-L activations contain extreme outliers that exceed the FP16 range and saturate} (Table~\ref{tab:overflow}); an engine containing an interval in which activations saturate returns a \emph{constant} regardless of its input.

\begin{table}[htbp]
\centering
\caption{Maximum absolute activation, against the FP16 ceiling of 65{,}504.}
\label{tab:overflow}
\begin{tabular}{lrl}
\toprule
Location & max.\ abs.\ value & vs.\ FP16 ceiling \\
\midrule
DINOv2-L, inside blocks 18--20 & 87 & fits \\
DINOv2-L, inside blocks 21--23 + head & \textbf{356{,}409} & $5.4\times$ over \\
DINOv3-L, stage boundaries (all resolutions) & \textbf{155{,}118} & $2.4\times$ over \\
\bottomrule
\end{tabular}
\end{table}

The locations of the outliers differ by model. In DINOv2-L only the last three blocks carry them, so FP32 on that interval alone suffices; in DINOv3-L all four parts must be FP32. We did not infer this from the boundary activations alone: we rebuilt each part in FP16 in turn and ran the resulting chain over all $n=1{,}882$ test images. All four produced output that was invariant to the input, which confirms the requirement directly. The broken chains also ran \emph{faster} than the correct one, so this check cannot be replaced by a timing comparison. This is the vision counterpart of the massive activations reported for language models~\cite{sun}, a known obstacle to quantization.

This failure is difficult to detect. The split ONNX graphs are bit-identical to the unsplit model, and dtypes, byte counts and memory formats are all correct. Moreover, the affected engine showed a shorter latency than the correctly configured one---the broken FP16 single engine of DINOv2-L runs in \SI{8.049}{\milli\second} against \SI{16.023}{\milli\second} for the correct FP32 build, so \emph{the faster engine is the broken one}; latency alone is therefore not evidence of correctness. The two builds differ in arithmetic precision and in kernel selection at the same time, so we do not attribute the speed difference to computation being elided---we did not separate its components. Three inexpensive checks detect this failure:

\begin{enumerate}
\itemsep0pt
\item verify that the output changes when the input changes;
\item inspect the maximum absolute activation \emph{inside} each interval, since outliers need not appear at a boundary;
\item compare engine sizes, since a part built in FP32 is about twice the size of the corresponding FP16 part.
\end{enumerate}

\subsection{Deployed configuration and measurements}
\label{sec:deploy}

With FP32 restricted to the intervals that carry outliers, DINOv2-L runs in \SI{8.657}{\milli\second} (5-way split, FP32 on the last part) and DINOv3-L in \SI{16.608}{\milli\second} (4-way split, all FP32), both at $N{=}32$. Classification accuracy does not fall below server FP32---DINOv2-L reaches 0.8443 (server 0.8438) and DINOv3-L 0.7476 (server 0.7471)---and argmax agreement is 99.79\% and 99.95\% at this resolution. Split correctness was verified twice. First, running the unsplit ONNX graph and the chained split graphs on identical inputs gives bit-identical outputs. Second, on the device, the split chain and the single-engine build agree on 99.79\% of the test set for DINOv2-L (1{,}878 of 1{,}882 images) and 99.89\% for DINOv3-L (1{,}880 images).

The preparation path must be held fixed across a resolution sweep: the optimisation that DINOv3-L requires rewrites the graph and changes how TensorRT fuses it, so mixing prepared and unprepared graphs across levels would misattribute a preparation effect to resolution. The deployed DINOv2-L configuration omits the optimisation, which it does not need.

\begin{figure}[t]
\centering
\includegraphics[width=\textwidth]{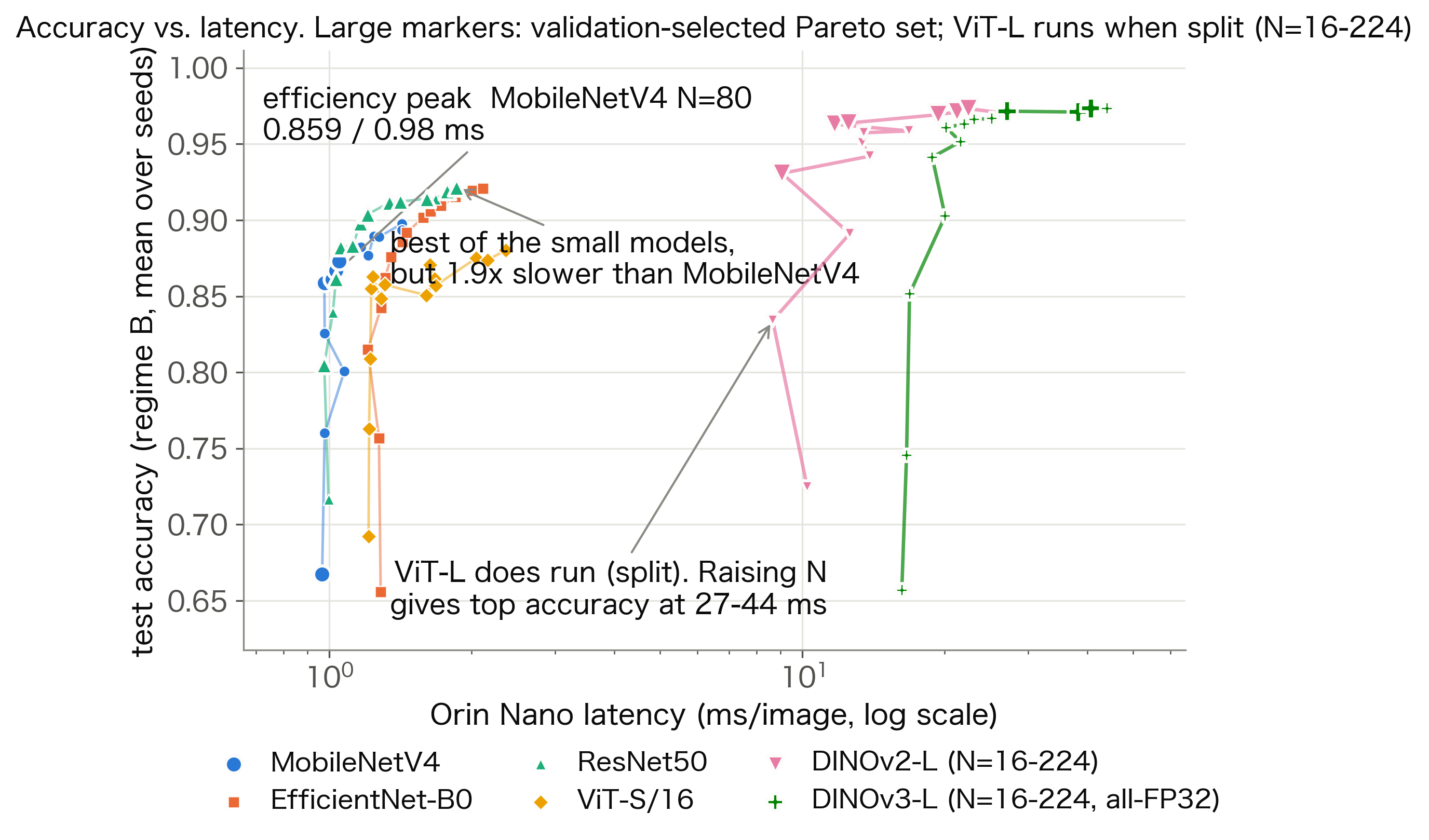}
\caption{Accuracy versus measured latency on Orin Nano. Large markers are the Pareto set selected on \emph{validation} over all 84 candidates, the same decision as the tables; because the vertical axis is test accuracy, a selected point can appear dominated there. The small models use FP16; DINOv2-L uses mixed FP16/FP32 and DINOv3-L uses FP32 throughout. The vertical axis is accuracy measured on the deployed Orin engines (30 seeds; 29 for the ViT-L-scale models).}
\label{fig:pareto}
\end{figure}

Figure~\ref{fig:pareto} places these points against the small models. ViT-L extends the Pareto front upward in accuracy, but at $N{=}112$ DINOv2-L costs $8.2\times$ the deployment-path classification time of MobileNetV4 ($12.1\times$ for DINOv3-L). Identifying one individual once after detection is feasible within a per-detection latency constraint; running every frame through ViT-L is not.

\subsection{Distinguishing agreement with the server from classification accuracy}
\label{sec:agreement}

Among the models that run \emph{entirely} in FP16---the CNNs and ViT-S/16---reduced-precision inference affects architectures very differently. The CNNs agree with server FP32 on 98.9--100\% of images, whereas ViT-S/16 alone falls to 90.65\% at $N{=}112$ and 86.08\% at $N{=}224$ (agreement rates are for a single seed, 42). At $N{=}16$ the same model still agrees on 99.73\%, so the failure is confined to the high-resolution end. An FP32 engine of the same ViT-S/16 model on the same device agrees with the server on 99.95\% of images; the loss is observed with the all-FP16 engine build used here. Rebuilt as an all-FP32 engine and measured over 30 seeds at all 14 resolutions, deployed validation returns to within $-0.07$ to $-0.01$ points of server FP32 (test: $-0.03$ to $+0.02$), at $2.5\times$ the FP16 GPU time (\S\ref{sec:optimal}).

The loss of agreement shows up as a loss of accuracy. Measuring all 56 CNN and ViT-S/16 configurations over 30 seeds, ViT-S/16 alone falls 4--7 percentage points below server FP32 at $N\ge96$ (from 0.9018 to 0.8576 at $N{=}112$, a loss of 4.42 points; from 0.9357 to 0.8801 at $N{=}224$, a loss of 5.56 points). At $N\le64$ the difference is within 0.1 points and at $N{=}80$ it is $-1.07$ points, so the large 4--7 point loss appears only from $N{=}96$ upward, with $N{=}80$ as the transition. The three CNNs stay within $\pm0.1$ points at all 14 resolutions, which places the effect of half precision within measurement noise for them. This is the difference that moves the selection away from ViT-S/16 at the targets 0.920--0.927 and 0.931--0.938 (\S\ref{sec:optimal}).

The ViT-L configurations behave in the opposite way. DINOv2-L (mixed FP16/FP32) agrees with the server on 97.9--99.8\% of images across all 14 resolutions and DINOv3-L (all FP32) on 99.9--100\%, with no monotone dependence on resolution (Table~\ref{tab:vitlres}). We also measured their \emph{task accuracy} on the deployed engines over 29 seeds: DINOv2-L falls $-0.17$ to $-0.00$ points below server FP32 and DINOv3-L $-0.02$ to $+0.01$, so the three-way ordering read off the agreement rates survives when accuracy itself is measured. Reduced precision is thus a problem for one specific configuration, ViT-S/16, and not for transformers in general nor for low resolution in general.

\begin{figure}[t]
\centering
\includegraphics[width=0.9\textwidth]{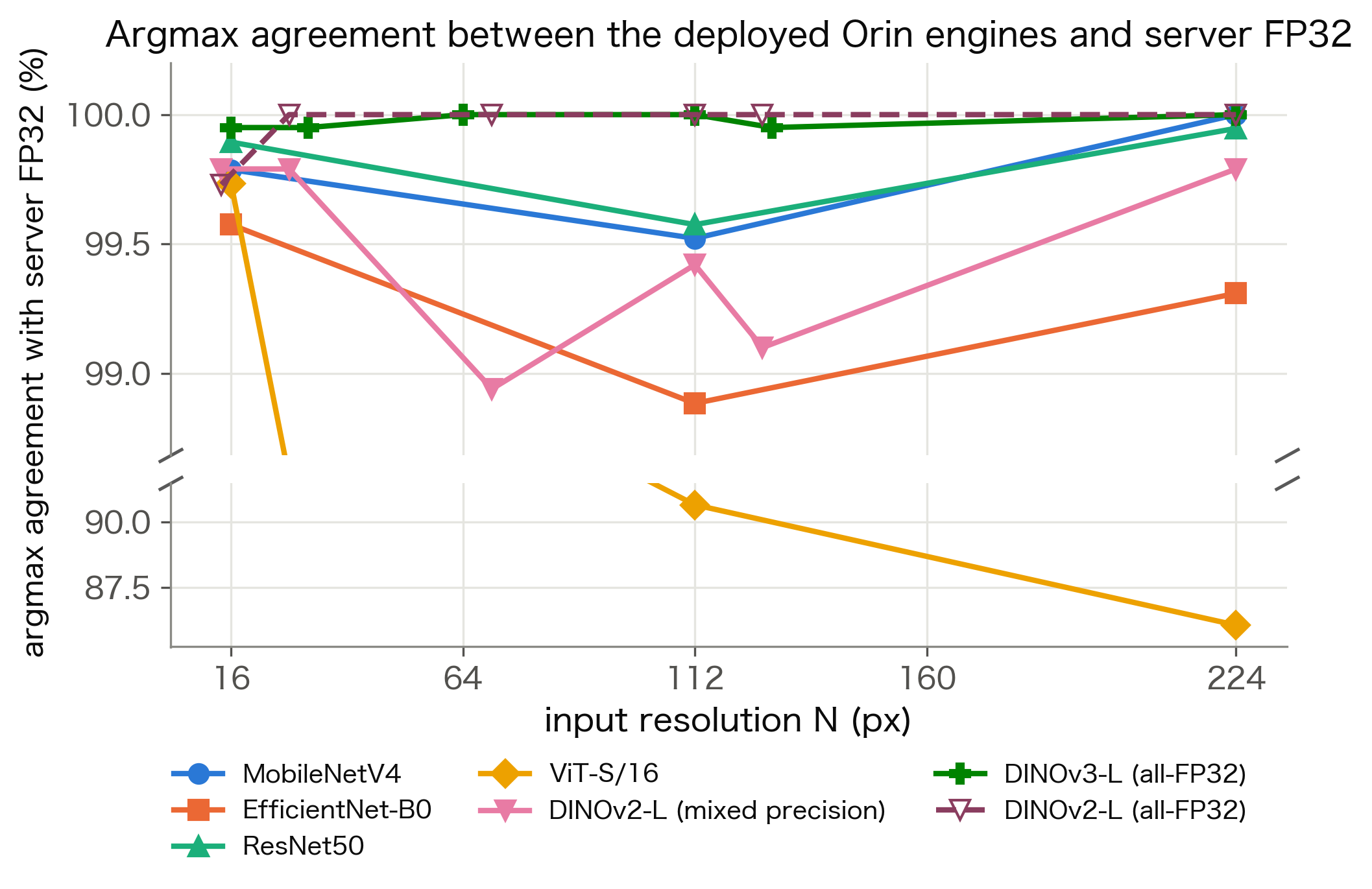}
\caption{Argmax agreement between the deployed Orin configurations and server FP32. The CNNs and ViT-S/16 run in FP16; DINOv3-L is FP32 throughout. For DINOv2-L both the deployed mixed-precision configuration (solid) and the all-FP32 control (dashed) are shown. The control increases agreement at five of the six tested resolutions, the largest increase being 1.06 percentage points at requested $N{=}64$; this indicates a small but measurable contribution from the precision configuration and does not establish low resolution as the cause of the remaining disagreement. ViT-L is measured at the six requested levels $N{=}16,32,64,112,128,224$; the horizontal axis is the side length actually fed to the model, so DINOv2-L appears at 14, 28, 70, 112, 126 and 224.}
\label{fig:fp16}
\end{figure}

To isolate the cause we ran a controlled comparison on DINOv2-L in which only the arithmetic precision was changed: weights, split points and preparation path were held identical to the deployed configuration (FP32 on the last part only), and every part was then rebuilt in FP32. The all-FP32 build agrees with the server on 99.7--100\% of images, within 1.1 percentage points of the deployed mixed-precision chain at all six tested resolutions ($N{=}16,32,64,112,128,224$; the control was not built at the other eight levels), which is why the dashed and solid curves in Figure~\ref{fig:fp16} lie close together. The control is nevertheless above the mixed-precision chain at five of those six levels, by up to 1.06 percentage points at requested $N{=}64$. Latency, meanwhile, rises from \SIrange{27.0}{62.4}{\milli\second} at $N{=}224$, a factor of 2.31. Mixed precision therefore costs almost nothing in reproducibility and less than half the time.

The practical implication is the same in both cases: agreement with a server reference and accuracy on the task are distinct quantities, and a deployment decision should evaluate them separately.

\section{Discussion}

\paragraph{Correspondence between design variables and their effects.} Replacing ViT-S/16 with MobileNetV4 at $N{=}224$ saves 40\% of GPU inference time (\SI{2.359}{\milli\second} to \SI{1.426}{\milli\second}), so the architecture is the more effective variable. The resolution is also effective: dropping MobileNetV4 from $N{=}224$ to $N{=}80$ saves 31\% of GPU inference time and 46.2\% of the classification time on the deployment path. On the offline evaluation path, the same change appears to save only 13.5\%, substantially understating the benefit on the deployment path.

\paragraph{Operational value of low-resolution robustness.} A model that holds accuracy at $N{=}48$--$80$ is both faster and cheaper to build optics for: the same camera identifies birds at longer range, or a less expensive lens covers the same range. It also keeps the specification inside what the atmosphere can deliver, which a high-$N$ design may not (\S\ref{sec:optics}). The required-resolution estimates provide initial targets for optical design, but they must be validated on representative PTZ imagery (\S\ref{sec:limits}).

\paragraph{When ViT-L is worth its cost.} ViT-L needs the fewest pixels \emph{and} costs the most time. That combination suits a two-stage pipeline: a cheap detector runs on every frame, and ViT-L identifies a detected individual once. At $N{=}128$ DINOv2-L costs \SI{16.8}{\milli\second} on the device at a measured on-device accuracy of 0.950 (Table~\ref{tab:vitlres}), which is compatible with a per-detection constraint, whereas per-frame use is not.

\paragraph{Practical checklist for half precision.} Inspect activation magnitudes \emph{inside} candidate intervals, not only at the boundaries; verify that engine outputs change when the input changes; and treat a build that is faster than expected as suspect. Engine size provides a simple diagnostic: a part that silently fell back to FP32 is about twice the size of the corresponding FP16 part.

\subsection{Distillation as a possible alternative to resolution-native retraining}
\label{sec:kd}

In \S\ref{sec:adapt} low-resolution accuracy was obtained by retraining at each resolution. Deployed lightweight models are often distilled instead~\cite{hinton}, so we asked whether distillation is an alternative lever, with everything but the loss held fixed: the student is EfficientNet-B0 trained at $N{=}224$, the teacher is our own DINOv2-L at $N{=}224$ (test accuracy 0.9772), and the loss is the standard form~\cite{hinton}, $0.5\,T^{2}\,\mathrm{KL}\!\left(\mathrm{softmax}(z_t/T)\,\|\,\mathrm{softmax}(z_s/T)\right) + 0.5\,\mathrm{CE}$ with $T=2$, where $z_s$ and $z_t$ are student and teacher logits. Data, splits, recipe and training budget are identical to the cross-entropy runs. This controlled experiment alone uses three seeds (42--44), not the 30 used elsewhere in the paper.

The results show no seed-consistent benefit at low resolution (Table~\ref{tab:kd}). For the six levels with $N\le96$, not one shows sign agreement across the three seeds, and the mean difference is $+0.14$ points---well inside the seed-to-seed standard deviation (1.5--4.3 points). A small seed-consistent benefit was observed only at $N{=}176$ and $N{=}224$ (mean $+0.36$ percentage points for $N \ge 112$). Distillation gives a small accuracy lift, but in this particular setting---one teacher, one student, one loss and temperature, and students trained at $N{=}224$ rather than natively at each resolution---we observed no consistent improvement in low-resolution robustness. This is an exploratory result on three seeds, not a general statement that distillation is ineffective, and it is not a full interaction study of native low-resolution training against distillation. Within the range covered here---one teacher--student pair, three seeds, students trained at $N{=}224$---the pipeline change of \S\ref{sec:adapt} moved low-resolution accuracy and this distillation setting did not, but that comparison is between two pipelines that differ in more than the loss, so it does not establish native retraining as the decisive mechanism.

\begin{table}[htbp]
\centering
\caption{Degradation robustness with and without distillation at selected resolutions (EfficientNet-B0, three seeds; accuracy on a 0--1 scale, differences in percentage points).}
\label{tab:kd}
\begin{tabular}{lrrr}
\toprule
$N$ & distilled & CE only & difference [pp] (sign agreement) \\
\midrule
16 & 0.2965 & 0.2779 & $+1.86$ (mixed) \\
32 & 0.4851 & 0.4853 & $\mathbf{-0.02}$ (mixed) \\
64 & 0.7558 & 0.7561 & $-0.04$ (mixed) \\
96 & 0.8574 & 0.8585 & $-0.11$ (mixed) \\
128 & 0.8978 & 0.8927 & $+0.51$ (mixed) \\
176 & 0.9145 & 0.9100 & $+0.44$ (\textbf{3/3}) \\
224 & 0.9242 & 0.9187 & $+0.55$ (\textbf{3/3}) \\
\bottomrule
\end{tabular}
\end{table}

\subsection{Effect of the power mode on speed and energy consumption}
\label{sec:power}

Because the board carries an on-board power monitor (INA3221), we logged supply power alongside latency. At idle, \texttt{VDD\_IN} draws \SI{3659}{\milli\watt} in \texttt{MAXN\_SUPER} against \SI{3570}{\milli\watt} in the \SI{15}{\watt} mode---a difference of 2.5\% relative to the \SI{15}{\watt} baseline.

\begin{figure}[t]
\centering
\includegraphics[width=\textwidth]{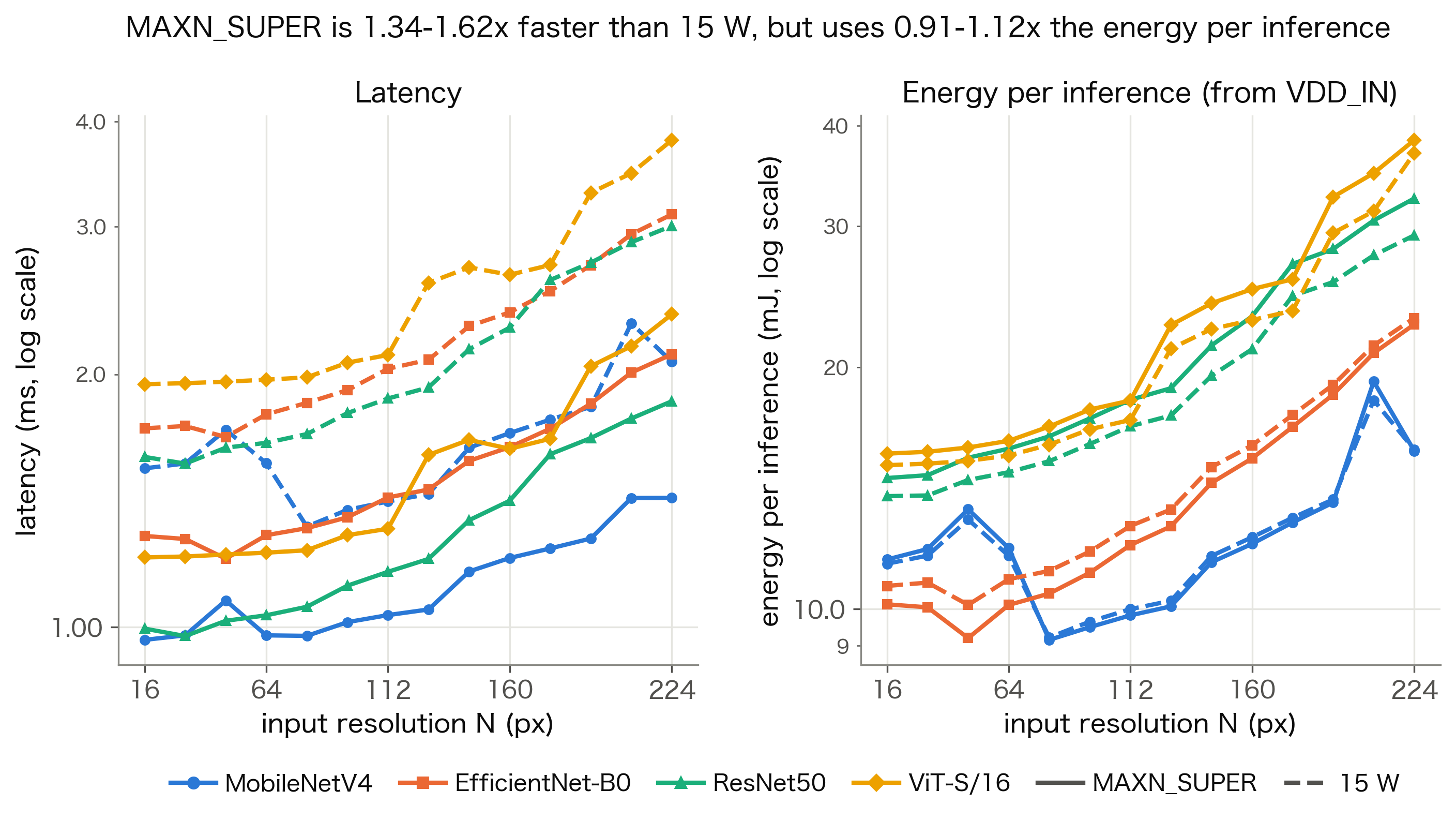}
\caption{\SI{15}{\watt} mode against \texttt{MAXN\_SUPER} for the four smaller models (56 configurations); the ViT-L split chains are summarised in the text. Left: latency in the two power modes. Right: energy per inference.}
\label{fig:power}
\end{figure}

Under load the behaviour differs. Over the 56 configurations of the CNNs and ViT-S/16, \texttt{MAXN\_SUPER} is a median $1.60\times$ faster than the \SI{15}{\watt} mode (range $1.34$--$1.62$), but it also draws a median $1.66\times$ the power ($1.27$--$1.80$). The energy per inference is therefore a median $1.03\times$ ($0.91$--$1.12$)---essentially unchanged (Figure~\ref{fig:power}). The 28 ViT-L split chains---measured under the same protocol but not shown in the figure---behave the same way: a median $1.59\times$ faster for a median $1.13\times$ the energy, i.e.\ slightly worse. \textbf{Raising the power mode shortens the wait but does not reduce the energy required to perform the same classification.} Where the installation runs on batteries or solar power, the decisive factor is the configuration choice of \S\ref{sec:optimal}, not the power mode.

\section{Limitations}

\label{sec:limits}
Every Orin latency reported here was measured with the device in its \texttt{MAXN\_SUPER} power mode; the \SI{15}{\watt} figures are used only for the comparison in \S\ref{sec:power}. Measurements were taken on an idle device at room temperature, so interference with a concurrently running detector was not evaluated. Our evaluation images are citizen-science photographs standing in for PTZ crops; flight pose, motion blur and outdoor illumination are not reproduced. TensorRT builds depend on measurement-driven kernel selection and are therefore not perfectly reproducible; whether a given part is built in FP16 or FP32 can vary between builds, which is detectable from the engine size. All reported Orin \emph{GPU benchmark} numbers are medians of 30 runs---only the H200 measurements rest on three runs---but a rebuild may select different kernels; the CPU preprocessing, single-classification wall-clock and integrated-throughput measurements follow the separate designs stated in \S\ref{sec:design} and where they appear. Four further limitations should be noted. The study uses a single dataset and a single primary task, so the required-resolution figures may not transfer to other species sets or imaging conditions. Only two devices were evaluated, one edge accelerator and one server GPU. Thirty seeds bound the precision with which small effects can be estimated; our claims are correspondingly restricted to effects whose direction is consistent across seeds within an architecture family. Interval estimates, and class-imbalance-aware metrics such as macro-F1 and per-class recall, are given for the central, selected configurations (\S\ref{sec:optimal}) but \emph{not for all 84 configurations}. And because the evaluation images are citizen-science photographs rather than PTZ crops, converting the required-resolution estimates into an optical specification requires additional validation. Both the reported test accuracies and the validation accuracies used to \emph{select} configurations are measured on the deployed engine for all six models; for the ViT-L-scale models this required rebuilding the engines deleted after the test measurement (\SI{27}{\giga\byte} per seed) and re-running all 29 seeds on the device (seed 42 is excluded because its ViT-L ONNX alone came from an older conversion path, so the CNNs rest on 30 seeds and the ViT-L-scale models on 29). With both sides aggregated over the same 29 seeds (43--71), the deployed validation means sit 0.02--0.24 points below server FP32 for DINOv2-L---the largest gap at $N{=}64$, the configuration selected under the 10 ms budget---and within $\pm0.01$ points for DINOv3-L, which is what moves the selection at targets 0.966--0.968, 0.972 and 0.975 and under the \SI{33.3}{\milli\second} budget. A related limitation concerns the split itself: our split is image-wise, so crops of the same source photograph, several photographs of the same observation and adjacent frames of the same video can straddle train and test. Counting groups by source-device ID, observation ID and video sequence, 61 test images (3.24\%) and 58 validation images (3.09\%) share a group with the training set, dominated by 51 adjacent video frames. We therefore built \emph{a sensitivity split that removes some forms of group sharing} (no group identified by that rule straddles; all 12{,}519 images retained; 70/15/15) and retrained and re-evaluated the 16 configurations at the selection boundary over 30 seeds in server FP32. Accuracy moves by $-0.74$ to $+1.00$ points in regime~A and $-1.21$ to $+0.86$ in regime~B, 30 of the 32 differences staying below one point; \textbf{all 14 non-trivial regime-difference signs are preserved} and only 2 of 120 pairwise orderings swap, both within 0.33 points and both involving ViT-S/16. These differences are not an estimate of how much the original split inflates accuracy through leakage: changing the split changes the training and evaluation samples themselves, so the two effects are not separated here. The check is also confined to server FP32 and to those 16 configurations: selection on the deployed engine, the Pareto set over all 84 candidates and the seed-stability analysis have \emph{not} been reproduced under the sensitivity split. The group labels are inferred from file names, so ``no group straddles'' holds for the groups that rule could identify. That limit bites once: the 1{,}000 frames extracted from video carry an eight-character video identifier at the head of the file name (33 videos) which the rule does not read, so each was treated as its own group. Counting by video instead, even this sensitivity split leaves 162 test images (8.6\%) and 161 validation images (8.6\%) sharing a video with the training set (under the image-wise split, 207 and 212), so it removes the sharing captured by the source-device, observation and video-sequence rule while leaving video-level sharing in place---which is why we do not call it group-disjoint. Within one video the consecutive retained frames are a median \SI{3.34}{\second} apart, with 278 of 967 consecutive pairs within \SI{1}{\second}. We therefore rebuilt the split under the video-aware rule (all 12{,}519 images retained; train 8{,}763, validation 1{,}880, test 1{,}876; no group under either rule straddles the splits) and retrained and re-evaluated the same 16 boundary configurations over 30 seeds in server FP32. Relative to the image-wise split, accuracy moves by $-1.45$ to $-0.21$ points in regime~A (all 16 negative) and $-1.66$ to $+0.02$ points in regime~B---a systematic decrease once video-level sharing is also removed---while all 14 non-trivial regime-difference signs are again preserved and 3 of 120 pairwise orderings swap, each within 0.61 points on the image-wise split. The same caveat applies: these differences are a sensitivity to the split, not an estimate of leakage-induced inflation. Per-model results are published as \texttt{summary\_group\_v2.json}, and both group rules are published with the manifest. Finally, the reported split-chain latency is the sum of the per-part GPU compute times; it includes the device-memory handover between parts but not host-side overhead. Two caveats attach to the deployment-path preprocessing figures in particular. They were measured under the assumption that the detector hands a decoded frame to the classifier, not on an integrated system. And decoding a 4K H.264 frame on the Orin CPU takes \SI{21.1}{\milli\second} at $3840\times1920$, which is 84\% of the \SI{25.0}{\milli\second} per-frame average that four streams at 10 fps allow (\S\ref{sec:design}). Using the hardware decoder (NVDEC) instead can shorten this substantially, but only if the frame is kept inside device memory: naively swapping the decoder makes things worse, because bringing the frame out to a BGR host array adds the NV12-to-BGRx and BGRx-to-BGR conversions and ends up slower than the CPU. Realising the gain requires the detector to consume device-memory buffers directly. A further caveat is that the two decoders do not agree pixel-wise---a systematic colour offset rather than rounding noise, most likely a BT.601/BT.709 or limited/full-range mismatch---so, since background subtraction thresholds on absolute intensity differences, switching decoders requires re-tuning the detection parameters. Following this route we integrated NVDEC into the detector and measured the \emph{detector-and-tracker pipeline}---decode, running-average background subtraction, contour extraction and tracking---running four 4K streams in four concurrent processes, \emph{without concurrent species classification}. Aggregate throughput is \SI{40.4}{fps} (median of 28 consecutive rounds; range \SI{39.1}{}--\SI{41.1}{fps}), so the minimum requirement of \SI{40}{fps} is met, but with almost no margin: only 20 of the 28 rounds reached \SI{40}{fps}. Pinning OpenCV's internal parallelism to a single thread was what closed the gap, raising per-process throughput from \SI{9.6}{}--\SI{9.9}{fps} to \SI{10.5}{}--\SI{11.0}{fps}; with four processes already competing for memory bandwidth, the thread pool costs more than it returns. The throughput of this pipeline while species classification runs on the same device remains to be evaluated, so the measurement is not a demonstration that detection and classification together sustain four streams at 10 fps. Sustained operation showed no thermal degradation: over 21.5 minutes the first 14 rounds averaged \SI{40.34}{fps} and the last 14 \SI{40.33}{fps}, while the enclosure temperature plateaued at \SI{71.4}{\celsius} (from \SI{58.8}{\celsius}) and the CPU held \SI{1.728}{\giga\hertz} in 241 of 245 samples.

\section{Conclusion}

We asked which input side length gives high identification accuracy at the shortest processing time, and answered it by combining a factorial sweep over 14 resolutions, six architectures, two training regimes and 30 seeds with on-device measurements. \textbf{The answer is conditional}: the configuration selected depends on the accuracy target, the candidate set, the preprocessing path and the precision assignment, and it is selected on validation and reported on test. Among the 84 base candidates, a validation target of 0.90 selects ResNet50 at $N{=}112$ (test 0.8980) at an estimated classification time of \SI{1.85}{\milli\second}, 0.93 selects ResNet50 at $N{=}224$ (test 0.9213, \SI{3.08}{\milli\second}) and 0.95 selects DINOv2-L at $N{=}144$ (test 0.9630, \SI{12.70}{\milli\second}); admitting 14 further ViT-S/16 FP32 configurations moves the recommendation over the 0.931--0.938 band and under the \SI{10}{\milli\second} budget. For the 0.90 target on the base candidate set the operating point is $N\approx112$, and raising $N$ beyond it yields little accuracy while increasing processing time.

Lowering $N$, by contrast, yields two benefits, the first of which is readily overlooked. On the deployment path---where the 4K frame is decoded once by the detector and the classifier only crops and resizes---going from $N{=}224$ to $N{=}80$ cuts the estimated classification time by 46.2\%. Measured instead on the offline evaluation path, in which every individual is decoded from its own file, the same change appears to save only 13.5\% and a floor of \SI{6.17}{\milli\second} appears to exist; that floor is an artefact of file-based evaluation and does not arise in deployment. The second benefit concerns the optics: the achievable range scales as $1/N$, and under the assumptions of \S\ref{sec:optics} atmospheric turbulence limits the number of independently resolvable elements at \SI{1}{\kilo\metre} to about 62, so a design premised on 224 \emph{independent} elements at that range is optimistic. That is a consideration derived from stated assumptions rather than a measurement of our system. When accuracy has to rise, changing the model gives the larger accuracy increment---though at several times the classification time, so it is not the cheaper lever---and it extends to ViT-L-scale models: they run on a Jetson Orin Nano even as a single unsplit engine, and splitting the graph at fusion boundaries confines FP32 to the intervals whose activations exceed the FP16 range, which makes the deployed chain $1.85\times$ faster than the unsplit build. Outlier-aware low-precision methods such as per-channel scaling may reduce the number of intervals that require FP32, which is the most direct route to making ViT-L-scale identification routine at the edge.

\section{Data and code availability}

\label{sec:availability}
The data manifest, both splits, every aggregate the tables and figures are built from, and the analysis, training and on-device measurement code are published at \url{https://doi.org/10.5281/zenodo.22697995} (concept DOI; mirrored at \url{https://github.com/focusnishikawa/bs2026-resolution-artifacts}).

\textbf{The images themselves are not redistributed}: iNaturalist photographs carry per-photograph licences chosen by their observers, Macaulay Library assets may not be redistributed, Wikimedia Commons files carry their own licences and attribution requirements, and the authors' own photographs and video frames are not released. What is published instead is one row per image---source, identifier, URL, SHA-256, group identifier under both rules, and membership in both splits. The manifest documents the provenance and split assignment of all 12{,}519 images and supports retrieval of the subset with available source links, under those sources' own terms. \textbf{The complete image set cannot currently be reconstructed from the public materials}, because some images are not released and others no longer carry a recoverable source URL. SHA-256 is given for all 12{,}519 images and 5{,}756 (46.0\%) carry a directly resolvable URL: iNaturalist observations (4{,}865) at \texttt{inaturalist.org/observations/<id>}, Wikimedia Commons files (746, reachable through the crawl log of visited URLs, since each file is named by the SHA-1 of its bytes), the BIRDS 525 Kaggle dataset (88) and Macaulay Library assets (57) at \texttt{macaulaylibrary.org/asset/<id>}. The remainder are the authors' own photographs (4{,}578 crops), their own video frames (1{,}000 with a video identifier and 339 without) and 846 images from earlier internal collections whose original URL is not recoverable.

Per-image predictions---six logits and the argmax for every image, for all seeds and configurations, 4{,}172 files---are attached to the Zenodo record rather than the repository.

Two levels of reuse should therefore be distinguished. \emph{Re-aggregation}---re-deriving every table, figure, selection rule and interval in this paper from the published aggregates and per-image predictions---requires no image and is supported in full. \emph{Reproducing the training} requires the images themselves, and is only partially supported: it is available for the 5{,}756 images with a resolvable source link, subject to each source's terms and to the possibility that a photograph has since been removed or relicensed, and not for the remainder.

\paragraph{Acknowledgements.} Computation was performed on the systems of the Foundation for Computational Science (FOCUS). The author used generative AI assistants (Claude, Anthropic; ChatGPT and Codex, OpenAI) for drafting and revising the manuscript, for analysis scripting and for internal manuscript review; all results, numbers and references were verified by the author.

\end{document}